\documentclass[accepted]{uai2021} 

\usepackage[american]{babel}

\usepackage{natbib} 
\usepackage{mathtools} 
\usepackage{booktabs} 
\usepackage{tikz} 

\renewcommand{\citet}[1]{\citeauthor{#1} (\citeyear{#1})}
\renewcommand{\citep}[1]{(\citeauthor{#1}, \citeyear{#1})}

\usepackage{xcolor}
\usepackage[normalem]{ulem}
\usepackage{algorithm}
\usepackage{algorithmic}
\usepackage{tikz}
\usetikzlibrary{bayesnet}
\usepackage{latexsym,amsfonts,amsmath,amssymb,mathrsfs,url,amsthm}
\usepackage{subcaption}
\usepackage{bbm}

\usepackage{bm}
\renewcommand{\boldsymbol}{\bm}

\newtheorem{theorem}{Theorem}

\newtheorem{proposition}{Proposition}

{
\theoremstyle{definition}

\newtheorem{remark}{Remark}
\newtheorem*{remark*}{Remark}

\newtheorem*{notation*}{Notation}

\newtheorem*{example*}{Example}

\newtheorem*{fact*}{Fact}
}

\newcommand{\rd}{\, \mathrm{d}}
\newcommand{\EE}{\mathbb{E}}
\newcommand{\PP}{\mathbb{P}}
\newcommand{\VV}{\mathbb{V}}

\newcommand{\KL}{\mathrm{KL}}
\newcommand{\card}{\mathrm{card}}

\newcommand{\bsone}{\boldsymbol{1}}
\newcommand{\Lcal}{\mathcal{L}}
\newcommand{\Lcalhat}{\hat{\mathcal{L}}}

\newcommand{\bsx}{\boldsymbol{x}}
\newcommand{\bsy}{\boldsymbol{y}}
\newcommand{\bsz}{\boldsymbol{z}}
\newcommand{\bsomega}{\boldsymbol{\omega}}
\newcommand{\bstheta}{\boldsymbol{\theta}}

\newcommand{\w}{w}
\newcommand{\bsw}{\boldsymbol{w}}
\newcommand{\Ocal}{\mathcal{O}}

\title{Efficient Debiased Evidence Estimation by\\ Multilevel Monte Carlo Sampling}

\author[1]{\href{mailto:Kei Ishikawa <kishikawa@ethz.ch>?Subject=Your UAI 2021 paper}{Kei~Ishikawa}} 
\author[2]{\href{mailto:Takashi Goda <goda@frcer.t.u-tokyo.ac.jp>?Subject=Your UAI 2021 paper}{Takashi~Goda}}
\affil[1]{%
    ETH Zurich\\
    Switzerland
}
\affil[2]{%
    The University of Tokyo\\
    Japan
}

\begin{document}
\maketitle


\begin{abstract}
  In this paper, we propose a new stochastic optimization algorithm for Bayesian inference based on multilevel Monte Carlo (MLMC) methods. In Bayesian statistics, biased estimators of the model evidence have been often used as stochastic objectives because the existing debiasing techniques are computationally costly to apply. To overcome this issue, we apply an MLMC sampling technique to construct low-variance unbiased estimators both for the model evidence and its gradient. In the theoretical analysis, we show that the computational cost required for our proposed MLMC estimator to estimate the model evidence or its gradient with a given accuracy is an order of magnitude smaller than those of the previously known estimators. Our numerical experiments confirm considerable computational savings compared to the conventional estimators. Combining our MLMC estimator with gradient-based stochastic optimization results in a new scalable, efficient, debiased inference algorithm for Bayesian statistical models.
\end{abstract}


\section{INTRODUCTION}

In empirical Bayes estimation, the model evidence or its lower bound are maximized to obtain a good estimate of the parameters. As such, the evidence maximization is considered a fundamental problem in Bayesian statistics and has been studied extensively for a long time. Perhaps the most common approach for the evidence maximization would be the expectation-maximization (EM) algorithm \citep{dempster1977maximum}. In the EM algorithm, the analytical form of the posterior distribution given data and parameters is required. In the case where the exact posterior distribution is not available, the algorithm can be extended by various approximation techniques such as variational EM algorithm \citep{jordan1999introduction} and Monte Carlo EM algorithm \citep{wei1990monte}. However, such approximation methods usually maximize a lower bound of the model evidence and thus the resulting estimates are biased.  

To reduce the bias from evidence estimation, application of the importance sampling \citep{robert2013monte} is often considered.
The estimate of the model evidence obtained by the importance sampling is known to have a negative bias, thus it serves as a stochastic lower bound of the true model evidence.
This stochastic lower bound can be maximized for the empirical Bayes method, by using its gradient with respect to the model parameters in stochastic optimization \citep{robbins1951stochastic}.  Especially, if the size of the data is too large to compute the gradient of the objective for all data points in each iteration, we can randomly pick a subset of the data to carry out doubly stochastic optimization. However, to reduce the bias of the objective, the number of the Monte Carlo samples required to compute the gradient for each data point needs to get larger, leading to a computational inefficiency in the optimization of the objective based on the importance sampling.

We tackle this problem by using a sophisticated Monte Carlo simulation technique called multilevel the Monte Carlo (MLMC) method. Although the MLMC was originally studied in the context of parametric integration \citep{heinrich1998monte} and stochastic differential equations \citep{giles2008multilevel}, it can be applied to many other contexts as well, in situations where the computational cost per Monte Carlo sample increases as we reduce the bias of an objective. By considering a hierarchy of different bias levels from a true objective and constructing a tightly coupled Monte Carlo estimator for the difference between two successive biased objectives, the true objective can be estimated quite efficiently compared to the standard Monte Carlo method which only estimates a single biased objective at a fixed bias level. In a favorable setting, the MLMC estimator can be even made unbiased for the true objective by some randomization \citep{rhee2015unbiased}. 

After the seminal work by \citet{giles2008multilevel}, the MLMC has been applied to various areas such as partial differential equations with random coefficients \citep{cliffe2011multilevel}, continuous-time Markov chains \citep{anderson2012multilevel} and Markov Chain Monte Carlo sampling \citep{dodwell2015hierarchical}. We refer the reader to a review on recent developments of the MLMC \citep{giles2015multilevel}.
Recently, the MLMC has been studied intensively for an efficient estimation of nested expectations, motivated by various applications such as computational finance \citep{bujok2015multilevel}, decision-making under uncertainty (\citeauthor{giles2020decision}, \citeyear{giles2020decision}; \citeauthor{hironaka2020multilevel}, \citeyear{hironaka2020multilevel}) and experimental design (\citeauthor{goda2020multilevel}, \citeyear{goda2020multilevel}; \citeauthor{goda2021design}, \citeyear{goda2021design}). 
In Appendix~\ref{app:review_MLMC}, we give a brief introduction to the MLMC method and its basic theory.

In this paper, the above-cited works on nested expectations are leveraged to obtain a computationally efficient, debiased estimator for the model evidence. We also provide an efficient, debiased estimator for the gradient of the model evidence with respect to the model parameters, which can be combined well with stochastic optimization to search for a good estimate of the parameters learned from a large data set.  

\section{BACKGROUND}

\subsection{Problem Settings}

\begin{figure}
    \centering
    \begin{tikzpicture}
  \node[obs]                (x) {$\bsx_n$};
  \node[latent, left=of x]  (z) {$\bsz_n$};
  \node[const, above=6mm of z] (t) {$\theta$};

  \edge {z} {x} ; %
  \edge {t} {x, z} ; %

  \plate {xz} {(x)(z)} {$n = 1,...,N$} ;
  
\end{tikzpicture}
\caption{A graphical model of Bayesian models with local latent variables.}
    \label{fig:graphical_model_1}
\end{figure}
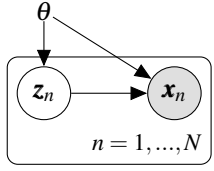

In this paper, we consider Bayesian statistical models with local latent variables that are formulated by the following i.i.d.\ data generating process:
\begin{equation}
\begin{aligned}
&\bsz_n &\sim& \ p_\theta(z), \\
&\bsx_n|\bsz_n=z_n &\sim& \ p_\theta(x|z_n),
\end{aligned}
\end{equation}
for $n=1,...,N$. Here, the bold letters $\bsx_n$ and $\bsz_n$ denote random variables, whereas the normal letter $x_n$ denotes the corresponding realization. This latent variable model can be expressed as a graphical model shown in Figure~\ref{fig:graphical_model_1}. The problem we are interested in is to estimate the parameter $\theta$ from the data $x_1, ..., x_N$.

Throughout the paper, we will omit the dependence of the model $p_\theta$ on the parameter $\theta$, and simply write $p$ instead of $p_\theta$ where it is obvious from the context. We also abbreviate a vector such as $(x_1, ..., x_n)$ as $x_{1:n}$ for the simplicity of notation.  To estimate the parameter $\theta$, we maximize the model evidence of the data $x_1, ..., x_N$, which is defined by 
\begin{align}
\Lcal(x_{1:N}; \theta)
&= \log p_\theta(x_{1:N}) \nonumber \\
&= \sum_{n=1}^N \log \int p_\theta(x_n| z_n) p_\theta(z_n)\rd z_n.\label{eq:log_likelihood}
\end{align}
Additionally, we assume that the size $N$ of the data is so large that stochastic or mini-batch optimization of the above objective is more desirable than batch optimization. Our aim of this paper is to introduce efficient, debiased Monte Carlo estimators of the model evidence \eqref{eq:log_likelihood} and its gradient with respect to $\theta$ and then to propose a new scalable, stochastic optimization algorithm of this objective.

\begin{remark}
Actually, it is possible to treat the parameter $\theta$ in a Bayesian manner and obtain a similar debiased, efficient estimator using MLMC. Bayesian treatment of parameters enables us to quantify the uncertainty of the estimated parameters. To calculate the debiased posterior, we can combine a gradient-based variant of the stochastic variational inference \citep{hoffman2013stochastic} with our proposed algorithm. Such a combination can be implemented very simply with almost no additional effort. The details on this point are discussed in Appendices~\ref{app:lmelbo} and \ref{app:lmelbo_exp}. 
\end{remark}

\subsection{Nested Monte Carlo Estimation of Model Evidence and its Gradient}

As discussed before, the model evidence can be estimated by the importance sampling. The estimator of the model evidence by the importance sampling, denoted here by $\Lcalhat_K$, is given by
$$\Lcalhat_K(x_{1:N}) = \sum_{n=1}^N \log\left[\frac{1}{K}\sum_{k=1}^{K}\frac{p(x_n| Z_{n, k})p(Z_{n,k})}{q_n(Z_{n,k})}\right],$$
where, for each $n$, $Z_{n,1}, ..., Z_{n, K}$ are i.i.d.\ random samples from a proposal distribution $q_n(z_n)$. In general, $q_n$ is taken so that it approximates the true posterior distribution of $\bsz_n$ given $\bsx_n=x_n$. This is because the Monte Carlo average inside the logarithm becomes exactly equal to the marginal distribution $p(x_n)$ with variance 0, if we can set $q_n(z_n)$ to the conditional distribution $p(z_n|\bsx_n = x_n)$. In practice, we choose the proposal distribution $q_n$ as an approximate posterior of $\bsz_n$ computed from $x_n$. So hereafter we will write $q(z_n; x_n)$ instead of $q_n(z_n)$ to express the proposal distribution of $Z_n$'s.

There are a few useful properties of this estimator for the model evidence based on the importance sampling, as proven in Theorem 1 of \citet{burda2015importance}. First, when we increase $K$, the number of the samples in the Monte Carlo average inside the logarithm, to infinity, we recover the model evidence thanks to the law of large numbers, i.e., we have
$$\lim_{K\to\infty}\Lcalhat_K(x_{1:N}) = \Lcal(x_{1:N}).$$
Second, when we denote the expectation of the estimator $\Lcalhat_K(x_{1:N})$ by $\Lcal_K = \EE[\Lcalhat_K(x_{1:N})]$, $\Lcal_K$ is always smaller than or equal to $\Lcal_{K+1}$. That is, we have
\begin{align}\label{eq:monotone}
\Lcal_1 \leq \cdots \leq \Lcal_K \leq \Lcal_{K+1} \leq \cdots \leq \Lcal_\infty = \Lcal(x_{1:N}),
\end{align}
which implies that the larger $K$ we use, the better lower bound on the model evidence we obtain.
Thus, the maximization of this lower bound with respect to $\theta$ is a good approximation of the evidence maximization if $K$ is chosen large enough.

In order to process a large data set efficiently, it is sensible to apply a gradient-based doubly stochastic optimization. For this, instead of looking at all the data points at each iteration, we randomly pick a subset (mini-batch) of the data points with size $M$, and estimate the corresponding log-marginal likelihood (or its gradient). This is equivalent to rewriting the model evidence \eqref{eq:log_likelihood} into a \emph{nested expectation}
\begin{align*}
\Lcal(x_{1:N}) = N\EE_{X}\left[ \log \int p_\theta(X| Z) p_\theta(Z)\rd Z\right],
\end{align*}
where $X$ is a random variable taking $x_1,\ldots,x_N$ uniformly and $\EE_X$ denotes the average with respect to $X$, and to estimate it by the nested Monte Carlo method:
\begin{align}
    \Lcalhat_{M,K} = \frac{N}{M}\sum_{m=1}^M \log\left[\frac{1}{K}\sum_{k=1}^{K}\frac{p(X_m| Z_{m, k})p(Z_{m,k})}{q(Z_{m,k};X_{m, k})}\right].\label{eq:nmc}
\end{align}
Here, it is important to note that $X_1,\ldots,X_M$ denote i.i.d.\ random samples of $X$. For each sample $X_m$, $Z_{m,1},\ldots,Z_{m,K}$ are conditionally i.i.d.\ random samples from a proposal distribution $q(z_m; X_m)$.
Because of the linearity of expectation, we have $\EE[\Lcalhat_{M,K}]=\EE[\Lcalhat_{1,K}]=\Lcal_{K}\leq \Lcal(x_{1:N})$.

Later in the paper, when we have only one sample of $X$, we write $\Lcalhat_{1, K}$ as $\Lcalhat_K := \Lcalhat_K(X_1, Z_{1,1:K})$ for notational simplicity.

The gradient of the model evidence with respect to $\theta$ can be estimated by the gradient of the nested Monte Carlo estimator \eqref{eq:nmc}, which is explicitly given by 
\begin{align*}
    \nabla_{\theta}\Lcalhat_{M,K} 
    & = \frac{N}{M}\sum_{m=1}^M \nabla_{\theta}\log\left[\frac{1}{K}\sum_{k=1}^{K}\frac{p(X_m| Z_{m, k})p(Z_{m, k})}{q(Z_{m, k};X_m)}\right] \\
    & = \frac{N}{M}\sum_{m=1}^M \frac{\displaystyle \frac{1}{K}\sum_{k=1}^{K}\frac{ \nabla_{\theta}\left(p(X_m| Z_{m, k})p(Z_{m, k})\right)}{q(Z_{m, k};X_m)}}{\displaystyle \frac{1}{K}\sum_{k=1}^{K}\frac{p(X_m| Z_{m, k})p(Z_{m, k})}{q(Z_{m, k};X_m)}}.
\end{align*}
Note that, by using automatic differentiation software, we can avoid coding the gradient in the presented form. We would emphasize here that  $\nabla_{\theta}\Lcalhat_{M,K}$ is a biased estimator of the gradient of the true model evidence, and so, applying a gradient-based doubly stochastic optimization based on $\nabla_{\theta}\Lcalhat_{M,K}$ does not converge to the optimal parameter $\theta$ in general.

Let us focus on the nested Monte Carlo estimation of the model evidence at this moment. A similar argument applies to the gradient of the model evidence. The mean squared error of $\Lcalhat_{M,K}$ is decomposed into the sum of the variance and the squared bias:
\begin{align*}
 & \EE\left[ \left( \Lcalhat_{M,K}-\Lcal(x_{1:N})\right)^2\right] \\
 & = \VV\left[\Lcalhat_{M,K}\right] + \left( \EE\left[\Lcalhat_{M,K}\right]-\Lcal(x_{1:N})\right)^2 \\
 & = \frac{\VV[\Lcalhat_{1,K}]}{M}+\left(\Lcal_K-\Lcal(x_{1:N})\right)^2.
\end{align*}
Therefore, in order to make the mean squared error small, we need to increase both the mini-batch size $M$ and the number of inner Monte Carlo samples $K$, so that the variance and the bias become small, respectively. Here, it follows from \eqref{eq:monotone} that the bias converges to 0 as $K$ approaches to infinity. More precisely, in order to estimate the model evidence with a mean squared accuracy $\varepsilon^2$, it suffices to have
\begin{align*} \frac{\VV[\Lcalhat_{1,K}]}{M}\leq \frac{\varepsilon^2}{2} \quad \text{and}\quad \left(\Lcal_K-\Lcal(x_{1:N})\right)^2\leq \frac{\varepsilon^2}{2}. \end{align*}
Assuming that $\VV[\Lcalhat_{1,K}]\approx \VV[\Lcalhat_{1, \infty}]$ and that the bias $|\Lcal_K-\Lcal(x_{1:N})|$ decays with the order $K^{-\alpha}$ for some $\alpha>0$, we need to set $M=O(\varepsilon^{-2})$ and $K=O(\varepsilon^{-1/\alpha})$, respectively. Since the computational cost of $\Lcalhat_{M,K}$ is given by the product $M\times K$, it is of $O(\varepsilon^{-2-1/\alpha})$. Although another balancing between the variance and the squared bias is possible, the cost of $O(\varepsilon^{-2-1/\alpha})$ cannot be improved by the nested Monte Carlo methods.

\subsection{Related Methods}

As discussed above, the nested Monte Carlo estimation is computationally inefficient. Thus there have been some attempts to improve the efficiency of the debiased estimation for the model evidence and its gradient. In the context of variational autoencoder (VAE) \citep{kingma2013auto}, the use of the nested Monte Carlo estimator has been actively studied \citep{burda2015importance} as it naturally extends the ELBO, the original objective of the VAE. As a more efficient variant of the nested Monte Carlo objective, applications of the Jackknife method and the Russian roulette estimator were proposed.

The Jackknife method is a bias removal method in statistics. It uses resampling techniques to remove low order bias, e.g., the bias of the first order Jackknife estimator becomes $O(n^{-2})$ as the $O(n^{-1})$ bias can be removed. \citet{nowozin2018debiasing} applied this idea to the estimation of the model evidence and its gradient of VAE.

More related to our work is a Russian roulette estimator introduced in \citet{luo2019sumo}. Their estimator, called SUMO, randomly picks a positive integer $\mathcal{K}$ from distribution 
\[    P(k \leq \mathcal{K}) = 
    \begin{cases}
    1/k &\text{if } k < a \\
    1/a \cdot (1 - 0.1)^{k-a} &\text{if } k \geq a,
    \end{cases} \]
for which $a=80$ is recommended in the paper. Here, the exponential decay of $\PP(k \leq \mathcal{K})$ for $k$ larger than $a$ serves as a soft truncation of the $\mathcal{K}$ as the $\mathcal{K}$ sufficiently larger than the $a$ cannot be sampled with high probability.  
This random sampling is applied to each data point $X$ and the SUMO outputs the following weighted sum:
\[ \Lcalhat_a^\text{SUMO}(X) = \sum_{k=1}^\infty\frac{\mathbbm{1}_{\{k \leq \mathcal{K}\}}}{\PP(k\leq\tilde{\mathcal{K}})}\widehat{[\Lcal_k - \Lcal_{k-1}]}(X),\]
where $\tilde{\mathcal{K}}$ is distributed identically to $\mathcal{K}$. The differences in the summation are defined by $\widehat{[\Lcal_K - \Lcal_{K-1}]}(X) = \Lcalhat_K(X;Z_{1:K}) - \Lcalhat_{K-1}(X; Z_{1:(K-1)})$, where we explicitly wrote the dependence of $\Lcalhat_K(x; Z_{1:K}) = \log\left[\frac{1}{K}\sum_{k=1}^{K}\frac{p(x| Z_k)p(Z_k)}{q(Z_k ;x)}\right]$ on $Z_{1:K}$ and $\Lcalhat_0$ is defined as $0$. 
The function $\mathbbm{1}_{\{k \leq \tilde{\mathcal{K}} \}}$ is the indicator function. Having defined the point-wise definition of the SUMO, the SUMO for mini-batch can simply defined as the following sum: $\Lcalhat_a^\text{SUMO}(X_{1:M}) = \sum_{m=1}^M  \Lcalhat_a^\text{SUMO}(X_m)$.


It should be also noted that this estimator is similar to the randomized MLMC estimator discussed in the next section, in that they both use estimators of differences with shared inner Monte Carlo samples. 
However, even though the SUMO attempted to construct an unbiased estimator of the model evidence, a truly unbiased estimation is infeasible in the sense that both the expected computational cost and the variance of the SUMO approach infinity as the bias tends to zero. 
Our estimator using the MLMC method, on the other hand, can completely remove the bias while requiring finite expected computational cost and having a bounded variance. We refer the reader to Appendix~\ref{app:comparison} for a detailed theoretical comparison between different estimators.

\section{PROPOSED ALGORITHM}

To reduce the necessary computational cost from $O(\varepsilon^{-2-1/\alpha})$ to $O(\varepsilon^{-2})$ for estimating the model evidence, we apply the MLMC methods. Later in this section, we also discuss the case for the gradient of the model evidence. 

The main difference from the nested Monte Carlo estimation is to consider a geometric hierarchy of the biased objectives $\Lcal_1, \Lcal_2,\ldots,\Lcal_{2^\ell},\ldots$ and to represent the model evidence by the following telescoping sum
\begin{align*}  
\Lcal(x_{1:N}) = \sum_{\ell=0}^{\infty}\left( \Lcal_{2^\ell}-\Lcal_{2^{\ell-1}}\right),
\end{align*}
where we set $\Lcal_{2^{-1}}\equiv 0$. We call the term with $\ell=0$ \emph{main term} and the remaining terms \emph{correction terms}. Note that truncating the infinite sum over $\ell$ up to the first $L$ terms yields the objective $\Lcal_{2^L}$. The nested Monte Carlo method estimates the single term $\Lcal_{2^L}$ only, whereas the MLMC method estimates the main term and the correction terms (up to level $L$) independently and sums them up. 

The key ingredient is in how to estimate the correction terms. In order to estimate $\Lcal_{2^\ell}$ by $\Lcalhat_{M,2^{\ell}}$ for some mini-batch size $M$, we generate $2^{\ell}$ i.i.d.\ samples of $Z_m$ from a proposal distribution $q(z_m; X_m)$ for $m=1,\ldots,M$. Here, for the same mini-batch, two halves of the i.i.d.\ samples of $Z_m$ can be used to compute $\Lcalhat_{M,2^{\ell-1}}$ twice, which we denote by $\Lcalhat_{M,2^{\ell-1}}^{(a)}$ and $\Lcalhat_{M,2^{\ell-1}}^{(b)}$, respectively. Defining
\[ \Delta \Lcalhat_{M,2^\ell} = \begin{cases} \Lcalhat_{M,1} & \text{if $\ell=0$,}\\
\displaystyle \Lcalhat_{M,2^{\ell}} - \frac{\Lcalhat_{M,2^{\ell-1}}^{(a)}+\Lcalhat_{M,2^{\ell-1}}^{(b)}}{2} & \text{otherwise,} \end{cases} \]
the linearity of expectation ensures that
\[ \EE\left[ \Delta \Lcalhat_{M,2^\ell} \right] = \Lcal_{2^\ell}-\Lcal_{2^{\ell-1}}. \]
This means that $\Delta \Lcalhat_{M,2^\ell}$ is an unbiased estimator for the correction term. Now the truncated telescoping sum is estimated by
\begin{align} \Lcalhat^{\text{MLMC}}_{2^L}:=\sum_{\ell=0}^{L}\Delta \Lcalhat_{M_\ell,2^\ell}, \label{eq:mlmc} \end{align}
for mini-batch sizes $M_0,\ldots,M_L>0$. This is our MLMC estimator for the model evidence, and we refer the readers to Algorithm \ref{algo:mlmc} for its summary.

\begin{algorithm}
\small
\caption{MLMC estimation of $\Lcal_{2^L}=\EE[\Lcalhat_{1,2^L}]$}\label{algo:mlmc}
\algsetup{indent=0.4cm}
\begin{algorithmic}[1]
\FOR{$m = 1,...,M_0$}
    \STATE sample $X_m$ randomly from $x_1,...,x_N$
    \STATE sample $Z_m \sim q(z_m;X_m)$
\ENDFOR
\vspace{0.5em}
\STATE $\Delta \Lcalhat_{M_0, 1} \leftarrow \frac{N}{M_0} \sum_{m=1}^{M_0} \log \left( \frac{p(X_m|Z_m)p(X_m)}{q(Z_m; X_m)} \right)$\\
\hskip\algorithmicindent \\
\FOR{$\ell = 1,...,L$}
    \FOR{$m = 1,...,M_\ell$}
        \STATE (re-)sample $X_m$ randomly from $x_1,...,x_N$
        \FOR{$k = 1,...,2^\ell$}
            \STATE (re-)sample $Z_{m, k} \sim q(z_m;X_m)$
        \ENDFOR
        \vspace{0.5em}
        \STATE $\Lcalhat^{(a)}_{2^{\ell-1}} \leftarrow N \cdot \log \left[ \frac{1}{2^{\ell-1}} \sum_{k=1}^{2^{\ell-1}} \frac{p(X_m|Z_{m, k})p(Z_{m, k})}{q(Z_{m, k}; X_m)}  \right]$ \\
        \vspace{0.5em}
        \STATE $\Lcalhat^{(b)}_{2^{\ell-1}} \leftarrow N \cdot \log \left[ \frac{1}{2^{\ell-1}} \sum_{k=2^{\ell-1}+1}^{2^\ell} \frac{p(X_m|Z_{m, k})p(Z_{m, k})}{q(Z_{m, k}; X_m)} \right]$ \\
        \vspace{0.5em}
        \STATE $\Lcalhat_{2^\ell} \leftarrow N \cdot \log \left[ \frac{1}{2^\ell} \sum_{k=1}^{2^\ell}  \frac{p(X_m|Z_{m, k})p(Z_{m, k})}{q(Z_{m, k}; X_m)}  \right]$ \\
        \vspace{0.5em}
        \STATE $\Delta_\ell \Lcalhat_m \leftarrow \Lcalhat_{2^\ell} - \frac{1}{2}\left( \Lcalhat^{(a)}_{2^{\ell-1}} + \Lcalhat^{(b)}_{2^{\ell-1}}\right)$
    \ENDFOR
    \vspace{0.5em}
    \STATE $\Delta \Lcalhat_{M_\ell, 2^\ell} \leftarrow \frac{1}{M_\ell} \sum_{m=1}^{M_\ell} \Delta_\ell \Lcalhat_m$
    \vspace{0.5em}
\ENDFOR\\
\hskip\algorithmicindent\\
\STATE $\Lcalhat^\text{MLMC}_{2^L} \leftarrow \sum_{\ell=0}^L \Delta \Lcalhat_{M_\ell, 2^\ell}$
\end{algorithmic}
\end{algorithm}

We give a heuristic explanation on why our MLMC estimator is more efficient than the nested Monte Carlo estimator. It is easy to see that the computational cost and the variance of \eqref{eq:mlmc} are given by
\[ \sum_{\ell=0}^{L}M_\ell 2^{\ell} \quad \text{and}\quad  \sum_{\ell=0}^{L}\frac{\VV[\Delta\Lcalhat_{1,2^{\ell}}]}{M_\ell}, \]
respectively. Because of the shared use of i.i.d.\ samples of $Z_m$ in computing $\Lcalhat_{M,2^{\ell-1}}^{(a)}$, $\Lcalhat_{M,2^{\ell-1}}^{(b)}$ and $\Lcalhat_{M,2^{\ell}}$, the difference $\Delta \Lcalhat_{M,2^\ell}$ is expected quite small in magnitude, particularly for large levels $\ell$. In fact, we can assume that the variance per one randomly chosen data point, i.e., $\VV[\Delta\Lcalhat_{1,2^{\ell}}]$, decays exponentially fast with respect to $\ell$:
\[  \VV[\Delta\Lcalhat_{1,1}] \gg \VV[\Delta\Lcalhat_{1,2}] \gg \cdots \gg \VV[\Delta\Lcalhat_{1,2^{\ell}}] \gg \cdots. \]
Thus, in order to estimate higher-level correction terms accurately so that
\[ \sum_{\ell=0}^{L}\frac{\VV[\Delta\Lcalhat_{1,2^{\ell}}]}{M_\ell} \leq \frac{\varepsilon^2}{2} \]
is satisfied, we can decrease mini-batch sizes $M_\ell$ exponentially fast:
\[  M_0 \gg M_1 \gg \cdots \gg M_\ell \gg \cdots. \]
This leads to a substantial saving of the required total computational cost as compared to the nested Monte Carlo method. 

Let us assume that $\VV[\Delta\Lcalhat_{1,2^{\ell}}]$ decays with the order of $2^{-\beta \ell}$ for some $\beta>0$. The method of Lagrange multipliers leads to an optimal allocation of mini-batch sizes $M_0,M_1,\ldots, M_L$ by minimizing the total cost with keeping the variance bounded by $\varepsilon^2/2$:
\[ \sum_{\ell=0}^{L}M_\ell 2^{\ell}+\lambda \left( \sum_{\ell=0}^{L}\frac{\VV[\Delta\Lcalhat_{1,2^{\ell}}]}{M_\ell} - \frac{\varepsilon^2}{2} \right). \]
It is an easy exercise to obtain
\[ M_{\ell} \propto \sqrt{\frac{\VV[\Delta\Lcalhat_{1,2^{\ell}}]}{2^{\ell}}}=O(2^{-(\beta+1)\ell/2}).\]
If $\beta>1$ holds, the terms with small $\ell$ are dominant in the sum $\sum_{\ell=0}^{L}M_\ell 2^{\ell}$, whereas, if $\beta<1$ holds, the terms with large $\ell$ are dominant. For the dividing case $\beta=1$, all the terms are approximately equal. 

\begin{remark}\label{rem:rmlmc}
For the case $\beta>1$, our MLMC estimator can be even made unbiased by applying a randomization technique from \citet{rhee2015unbiased}. For any sequence $\bsomega=(\omega_0,\omega_1,\ldots)$ such that $\omega_{\ell}>0$ and $\|\bsomega\|_1=1$, it is possible to represent the model evidence by the weighted telescoping sum 
\begin{align*}  
\Lcal(x_{1:N}) = \sum_{\ell=0}^{\infty}\omega_{\ell} \frac{ \Lcal_{2^\ell}-\Lcal_{2^{\ell-1}}}{\omega_{\ell}} = \sum_{\ell=0}^{\infty}\omega_{\ell} \frac{ \EE\left[ \Delta \Lcalhat_{1,2^\ell} \right]}{\omega_{\ell}},
\end{align*}
For a mini-batch size $M>0$, let $\ell^{(1)},\ldots,\ell^{(M)}\geq 0$ be i.i.d.\ random samples from a discrete distribution with probabilities $\omega_0,\omega_1,\ldots$. Then the \emph{randomized} MLMC estimator
\[ \frac{1}{M}\sum_{m=1}^{M}\frac{\Delta \Lcalhat_{1,2^{\ell^{(m)}}}}{\omega_{\ell^{(m)}}}\]
becomes an unbiased estimator of $\Lcal(x_{1:N})$. The expected computational cost and the variance per one data point from the mini-batch are given by
\[ \sum_{\ell=0}^{\infty}\omega_{\ell}2^{\ell}\quad \text{and}\quad  \sum_{\ell=0}^{\infty}\frac{\VV[\Delta \Lcalhat_{1,2^{\ell}}]}{\omega_{\ell}}, \]
respectively. In order for these quantities to be both finite, it suffices to set $\omega_{\ell}\propto 2^{-(\beta+1)\ell/2}$. Such a discrete probability distribution does not exist if $\beta\leq 1$.
\end{remark}

We now come to estimation of the gradient of the model evidence. Similarly to the model evidence itself, we represent the gradient by the telescoping sum
\[ \nabla_{\theta}\Lcal(x_{1:N}) = \sum_{\ell=0}^{\infty}\left( \nabla_{\theta}\Lcal_{2^\ell}-\nabla_{\theta}\Lcal_{2^{\ell-1}}\right). \]
The correction terms can be estimated by
\[ \nabla_{\theta}\Delta \Lcalhat_{M,2^\ell} = \nabla_{\theta}\Lcalhat_{M,2^{\ell}} - \frac{\nabla_{\theta}\Lcalhat_{M,2^{\ell-1}}^{(a)}+\nabla_{\theta}\Lcalhat_{M,2^{\ell-1}}^{(b)}}{2}, \]
which is unbiased. In this way the truncated telescoping sum for the gradient is estimated by
\[ \nabla_{\theta}\Lcalhat^{\text{MLMC}}_{2^L}:=\sum_{\ell=0}^{L}\nabla_{\theta}\Delta \Lcalhat_{M_\ell,2^\ell}, \]
for mini-batch sizes $M_0,\ldots,M_L>0$. This is our MLMC estimator for the gradient of the model evidence. By a reasoning similar to before, we can expect a situation with
\[  \VV[\nabla_{\theta}\Delta\Lcalhat_{1,1}] \gg \cdots \gg \VV[\nabla_{\theta}\Delta\Lcalhat_{1,2^{\ell}}] \gg \cdots, \]
which allows for a rapid decrease
\[  M_0 \gg \cdots \gg M_\ell \gg \cdots, \]
resulting in a substantial computational saving as compared to the nested Monte Carlo estimator. Although it is not necessarily the case that $\VV[\nabla_{\theta}\Delta\Lcalhat_{1,2^{\ell}}]$ decays with the order of $2^{-\beta \ell}$ for the same $\beta$ appearing in the decay of  $\VV[\Delta\Lcalhat_{1,2^{\ell}}]$, the theoretical results given in the next section state under some assumptions that we have $\beta=2$ for both the model evidence and its gradient. Therefore, by following Remark~\ref{rem:rmlmc}, the gradient of the model evidence can be  estimated by the randomized MLMC method without any bias. 

\section{THEORETICAL RESULTS}\label{sec:theory}

In order to show that the necessary computational cost for our MLMC estimator of the model evidence to achieve a mean squared accuracy $\varepsilon^2$ is of $O(\varepsilon^{-2})$, we need to introduce the fundamental theorem on MLMC methods proven by \citet{giles2008multilevel} and \citet{cliffe2011multilevel}. Although a general statement is given in Appendix~\ref{app:review_MLMC} (Theorem~\ref{thm:mlmc_general}), the statement below is adapted for the current context. 
\begin{theorem}\label{thm:mlmc}
Assume that there exist positive constants $c_1,c_2,\alpha,\beta$ such that  
\begin{enumerate}
    \item $\alpha\geq \min(\beta,1)/2$,
    \item $|\Lcal_{2^{\ell}}-\Lcal(x_{1:N})|\leq c_12^{-\alpha \ell}$, and 
    \item $\VV[\Delta\Lcalhat_{1,2^{\ell}}] \leq c_22^{-\beta \ell}$.
\end{enumerate}
Then, for any given accuracy $\varepsilon<\exp(-1)$, there exists a positive constant $c_3$ such that there are the corresponding maximum level $L$ and the mini-batch sizes $M_0,M_1,\ldots, M_L$ for which the mean squared error of the MLMC estimator $\Lcalhat^{\text{MLMC}}_{2^L}$ is less than $\varepsilon^2$ with the total computational cost $C$ bounded by
\[ \EE[C]\leq \begin{cases} c_3\varepsilon^{-2}, & \beta>1, \\ c_3\varepsilon^{-2}|\log \varepsilon^{-1}|^2, & \beta=1, \\ c_3\varepsilon^{-2-(1-\beta)/\alpha}, & \beta<1. \end{cases} \]
\end{theorem}

\begin{remark}\label{rem:mlmc}
If $\beta>1$, the MLMC estimator can achieve the optimal computational complexity $O(\varepsilon^{-2})$ to estimate the model evidence. Notably, even if $\beta<1$, the cost of order $\varepsilon^{-2-(1-\beta)/\alpha}$ still compares favorably with the nested Monte Carlo estimator for which the cost is of $O(\varepsilon^{-2-1/\alpha})$.
\end{remark}

\begin{remark}\label{rem:multivariate_case}
A statement similar to Theorem~\ref{thm:mlmc} also holds for the gradient of the model evidence. The difference is that, since the gradient is represented as a vector, the second and third assumptions should be replaced, respectively, by
\begin{enumerate}
  \setcounter{enumi}{1}
  \item $\|\nabla_{\theta}\Lcal_{2^{\ell}}-\nabla_{\theta}\Lcal(x_{1:N})\|_2\leq c_12^{-\alpha \ell}$, and
  \item $\EE\left[\|\nabla_{\theta}\Delta\Lcalhat_{1,2^{\ell}}-\EE[\nabla_{\theta}\Delta\Lcalhat_{1,2^{\ell}}]\|_2^2\right] \leq c_22^{-\beta \ell}$,
\end{enumerate}
and that the mean squared error is given by the sum of the mean squared errors over all the elements. We refer to Section~2.5 of \citet{giles2015multilevel} for an extension of the MLMC theory to multi-dimensional outputs.
\end{remark}

Based on Theorem~\ref{thm:mlmc}, it suffices to characterize the values of $\alpha$ and $\beta$ for the MLMC estimator. The following result for the model evidence is an immediate consequence from \citet{goda2020multilevel}. 
We show a full proof in Appendix~\ref{app:proofs} for the sake of completeness.
\begin{theorem}\label{thm:evidence}
If there exist $s,t>2$ with $(s-2)(t-2)\geq 4$ such that
\begin{align*} 
& \EE_{X}\left[ \int \left|\frac{p_\theta(X| Z) p_\theta(Z)}{p_{\theta}(X)q(Z;X)}\right|^s\rd Z\right]<\infty, \quad\text{and}\\
& \EE_{X}\left[ \int \left|\log \frac{p_\theta(X| Z) p_\theta(Z)}{p_{\theta}(X)q(Z;X)}\right|^t\rd Z\right]<\infty,
\end{align*}
the MLMC estimator for the model evidence satisfies
\[ \alpha = \min\left\{\frac{s(t-1)}{2t}, 1 \right\}\; \; \text{and}\; \; \beta=\min\left\{\frac{s(t-2)}{2t}, 2\right\}. \]
\end{theorem}

It is important to see that $\beta\geq 1$ whenever $(s-2)(t-2)\geq 4$, which directly implies that the MLMC estimator achieves the optimal order of computational cost. Moreover, if $(s-4)(t-2)\geq 8$, we have $\beta=2$.

Regarding the gradient of the model evidence, we need to extend the result from the work by \citet{hironaka2020multilevel} which has been studied in a different context and only dealt with a scalar output instead of vector. The following result is an analogy of the one shown in \citet{goda2021design}. A full proof is given in Appendix~\ref{app:proofs}.
\begin{theorem}\label{thm:gradient}
If there exists $s\geq 2$ such that
\begin{align*} 
& \EE_{X}\left[ \int \left|\frac{p_\theta(X| Z) p_\theta(Z)}{p_{\theta}(X)q(Z;X)}\right|^s\rd Z\right]<\infty, \quad\text{and}\\
& \sup_{x,z}\left\|\nabla_{\theta}\log p_\theta(x| z) p_\theta(z) \right\|_{\infty}<\infty,
\end{align*}
the MLMC estimator for the gradient of the model evidence satisfies
\[ \alpha = \min\{s/2, 1\} \quad\text{and}\quad \beta = \min\{ s/2, 2\}. \]
\end{theorem}

Again we see that $\beta\geq 1$, which directly implies that the MLMC gradient estimator achieves the optimal order of computational cost in Theorem \ref{thm:mlmc}. Moreover, if $s\geq 4$, we have $\beta=2$. Therefore, the assumptions made in Theorems~\ref{thm:evidence} and \ref{thm:gradient} are satisfied simultaneously for large $s$ and $t$, the MLMC estimators for the model evidence and its gradient both attain $\beta=2$, which will be supported by the numerical results given in the next section.

\section{EXPERIMENTS}

To illustrate the effectiveness of our MLMC approach, we compared the computational efficiency of several evidence estimation methods using a random effect logistic regression model. In Appendix~\ref{app:lmelbo_exp}, we additionally provide experimental results of Bayesian version of random effect logistic regression and (sparse) Gaussian process classification. In all experiments, our algorithm was run on a single CPU, and Python codes used in our experiment are available at 
\url{https:\\github.com/Goda-Research-Group/mlmc-model-evidence}.

\subsection{Experimental Settings}

The random effect logistic regression is a model of the following i.i.d.\ data generating process for $n = 1, 2, \ldots, N$ and $t=1, \ldots, T$:
\begin{align*}
    \bsz_n &\sim N(0,\tau^2) \\
    \bsy_{n,t} &\sim \text{Bernoulli}\left(p_n\right),
\end{align*}
where we set the logit $p_n$ to $p_n = \sigma(\bsz_n + \w_0 + \w^T x_{n,t})$ for the sigmoid link function $\sigma(x)=1/(1 + \exp(-x))$. 
This model explains the binary response $\bsy_{n, t}$ of each individual $n$ at each time point $t$ given an explanatory variable $x_{n, t}$. By adding a random effect term $\bsz_{n}$ to the simple logistic regression model, we can estimate the effect $(\w_0, \w)$ of $x_{n,t}$ on $\bsy_{n, t}$ more accurately by removing the individual variations in the data. 

In our experiment, we used a synthetic data generated from a model whose parameters are given by $\eta=1.0$, $\w_0=0$, $\w=(0.25, 0.50, 0.75)^T$. The explanatory variables $x_{n, t}$'s are all taken from a standard normal distribution and $T$ was set to 2. To keep the parameter $\tau^2$ positive, we parametrized it with a non-constrained parameter $\alpha$ by the softplus transformation as $\tau^2=\log{\left(1+\exp(\eta)\right)}$. For choosing a proposal distribution $q(z_n; x_{n, 1:T})$, we used the Laplace approximation \citep{bishop2006pattern}. For the optimization, the Adam optimizer \citep{kingma2014adam} was used.

\subsection{Convergence of MLMC Coupling}
To examine whether the assumptions required for the MLMC estimation in Theorem~\ref{thm:mlmc} are satisfied, we evaluated the convergence behavior of the corrections $\Delta\Lcalhat_{1,2^{\ell}}$ and their gradient counterparts $\nabla_{\theta}\Delta\Lcalhat_{1,2^{\ell}}$.  

\begin{figure}
    \centering
    \begin{subfigure}{0.23\textwidth}
    \hspace*{-2em}                                                           
    \includegraphics[width=1.25\linewidth]{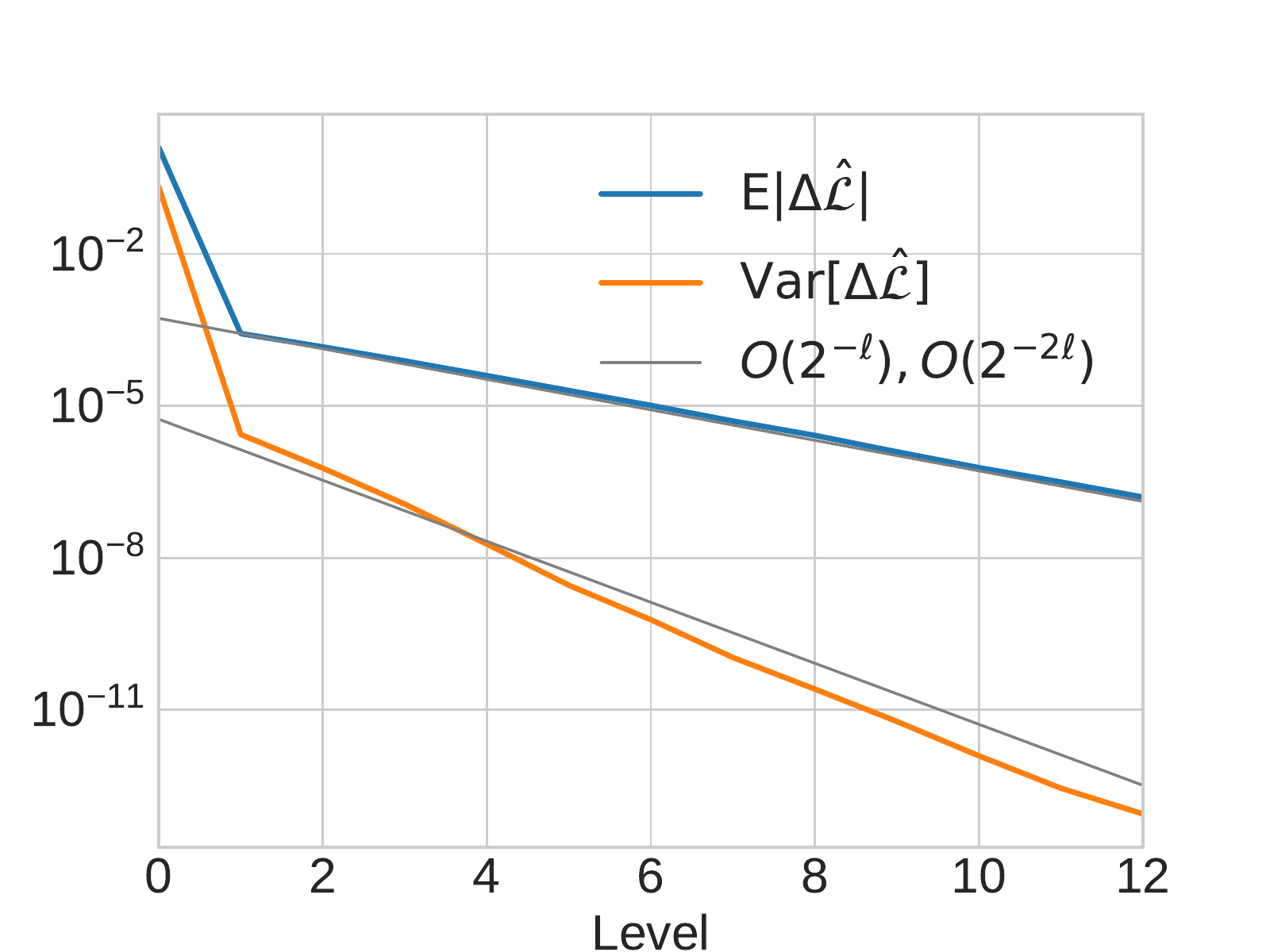}
    \caption{Decay of $\Delta\Lcalhat_{1,2^{\ell}}$}
    \label{fig:conv-delta} 
    \end{subfigure}
    \begin{subfigure}{0.23\textwidth}
    \hspace*{-1em}                                                           
    \includegraphics[width=1.25\linewidth]{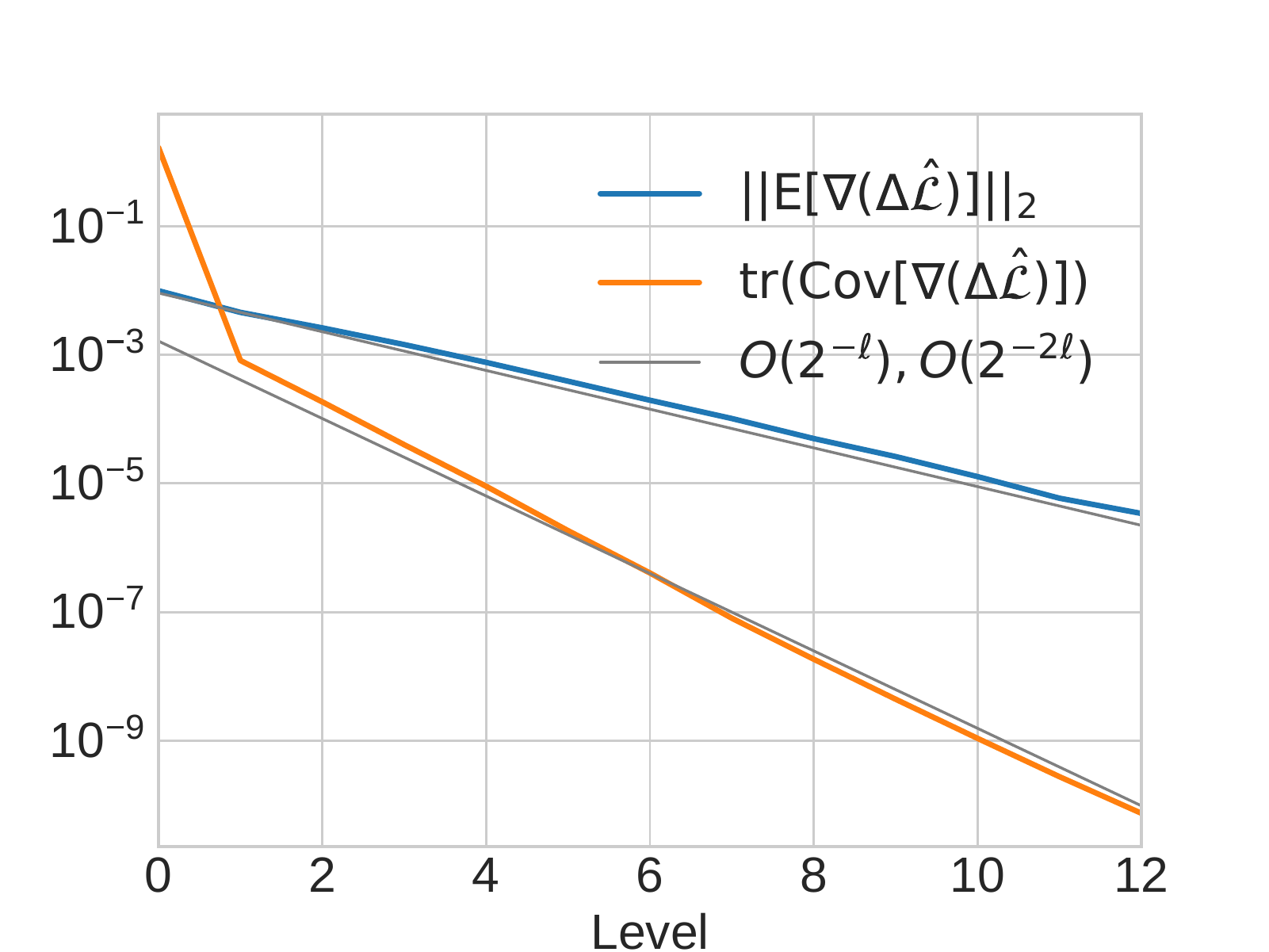}
    \caption{Decay of $\nabla_{\theta}\Delta\Lcalhat_{1,2^{\ell}}$}
    \label{fig:conv-grad-delta} 
    \end{subfigure}
    \caption{Convergence of the mean and variance of the coupled correction estimators.}
\end{figure}

Figure~\ref{fig:conv-delta} shows the convergence behaviors of $\EE[\Delta\Lcalhat_{1,2^{\ell}}]$ and $\VV[\Delta\Lcalhat_{1,2^{\ell}}]$. We see that  $\EE[\Delta\Lcalhat_{1,2^{\ell}}]$ and $\VV[\Delta\Lcalhat_{1,2^{\ell}}]$ approximately decay with the orders of $2^{-\ell}$ and $2^{-2\ell}$, respectively, implying that we have $\alpha=1$ and $\beta=2$ in the assumptions of Theorem~\ref{thm:mlmc}. 

Figure~\ref{fig:conv-grad-delta} shows the convergence behaviors of $\EE[\nabla_{\theta}\Delta\Lcalhat_{1,2^{\ell}}]$ and $\mathrm{Cov}[\nabla_{\theta}\Delta\Lcalhat_{1,2^{\ell}}]$ in $L_2$ norm and in trace, respectively. The trace of the covariance is equivalent to $\EE\left[\|\nabla_{\theta}\Delta\Lcalhat_{1,2^{\ell}}-\EE[\nabla_{\theta}\Delta\Lcalhat_{1,2^{\ell}}]\|_2^2\right]$ appearing in Remark \ref{rem:multivariate_case}. Again, the requirements for the MLMC method, i.e., the exponential decays of the corrections terms, are satisfied for the mean and the variance of the gradient counterparts. These numerical results support the theoretical findings given in Section~\ref{sec:theory}.

\subsection{Accuracy of Estimation by MLMC}

\begin{table*}
\small
\begin{center}
\caption{Accuracy of Parameter Estimation by Different Objectives and Estimation Methods. }
\small

\begin{tabular}{ l  r  r  r  r  r  r } 
\toprule
 & $\eta$  & $\w_0$  & $\w_1$  & $\w_2$  & $\w_3$  & MSE \\
\midrule
Ground Truth & 1.0 & 0.0 & 0.25 & 0.5 & 0.75 & 0.0 \\
NMC (K=1)     & -0.272 ± 0.121 & -0.003 ± 0.023 & 0.231 ± 0.021 & 0.456 ± 0.021 & 0.684 ± 0.022 & 1.6412 \\
NMC (K=8)     &  0.546 ± 0.086 & -0.005 ± 0.023 & 0.244 ± 0.021 & 0.485 ± 0.020 & 0.712 ± 0.018 & 0.2167 \\
NMC (K=64)    &  0.894 ± 0.059 &  \textbf{0.002} ± 0.024 & 0.252 ± 0.019 & 0.480 ± 0.021 & 0.744 ± 0.022 & 0.0169 \\
NMC (K=512)   &  1.038 ± 0.049 &  0.012 ± 0.021 & 0.244 ± 0.021 & \textbf{0.496} ± 0.020 & \textbf{0.747} ± 0.022 & 0.0059 \\
MLMC (L=9)    &  1.052 ± 0.033 &  0.010 ± 0.006 & \textbf{0.250} ± 0.005 & 0.511 ± 0.003 & 0.741 ± 0.003 & 0.0041 \\
RandMLMC (L=9)&  \textbf{0.966} ± 0.033 & -0.003 ± 0.006 & 0.241 ± 0.004 & 0.507 ± 0.003 & 0.744 ± 0.003 & \textbf{0.0026} \\
SUMO (K=512)  &  0.951 ± 0.083 & -0.011 ± 0.013 & 0.242 ± 0.008 & 0.506 ± 0.009 & 0.739 ± 0.009 & 0.0101 \\
Jackknife (K=512)   &  0.959 ± 0.053 & -0.016 ± 0.020 & 0.248 ± 0.021 & 0.494 ± 0.022 & 0.745 ± 0.016 & 0.0065 \\

\bottomrule
\end{tabular}
\label{tab:accuracy}
\end{center}
\end{table*}

Next, we compared the estimation accuracy of several evidence estimation methods by changing $K$'s for the biased objective, $\Lcal_K$. In Table~\ref{tab:accuracy}, the means and standard deviations of the estimated parameters obtained from different objectives are listed. For each method and objective, the parameters were estimated 100 times to obtain these quantities. The soft truncation of the SUMO was replaced by the hard truncation, to match the bias of all estimators. We can see that the biases become smaller as we increase $K$ and the smallest bias is attained for $K=512$ (or $L=9$). When we look at the standard deviations of the estimates, both randomized and non-randomized MLMC methods yield smaller standard deviations and mean squared errors (MSEs) than other methods with the same bias size. This is because the gradient estimation by the MLMC method has a smaller variance than other methods.  

\begin{figure}
    \centering
    \includegraphics[width=\linewidth, trim=0mm 5mm 0mm 10mm, clip]{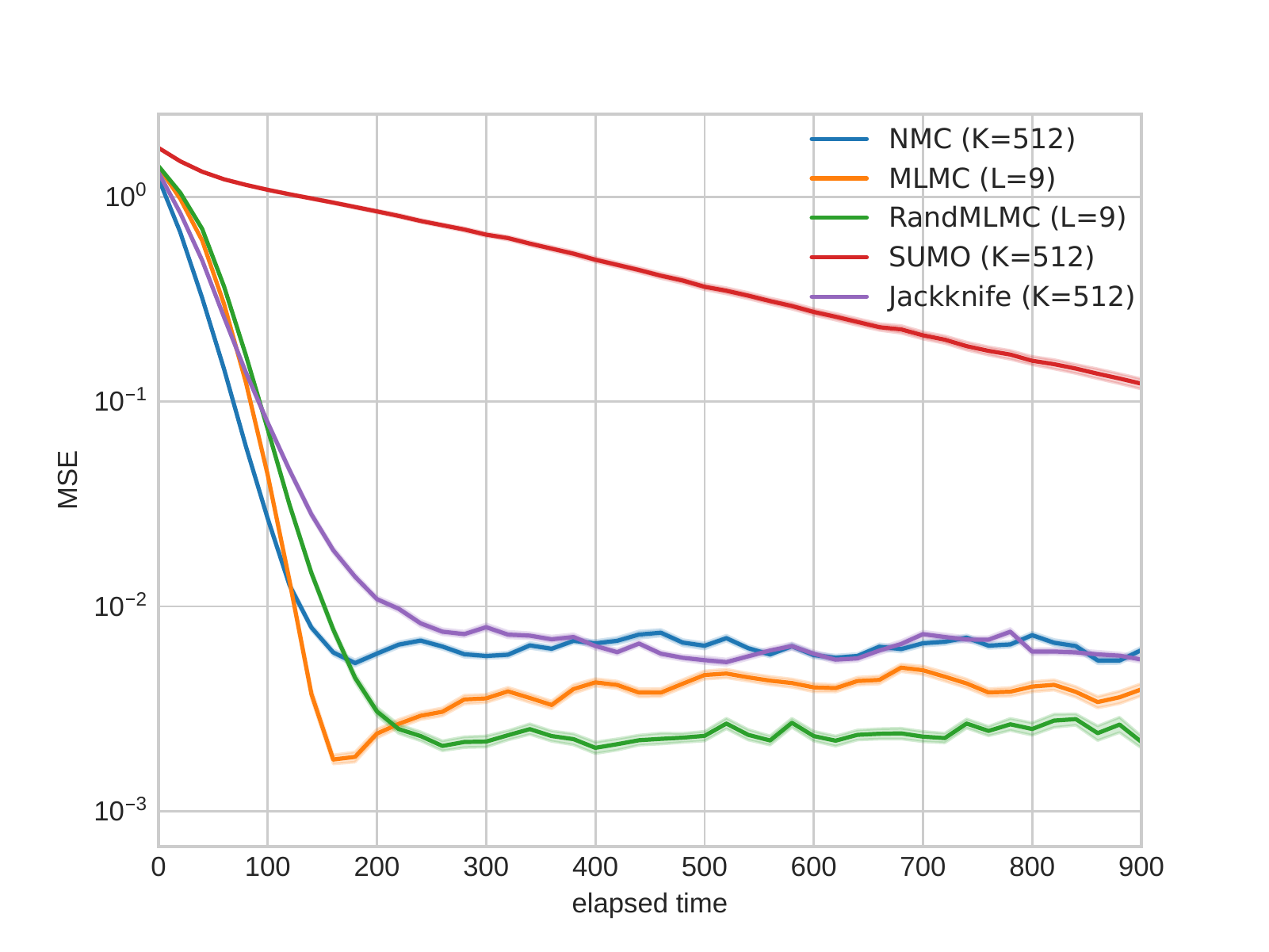}
    \caption{Learning curves of the different objective functions. Instead of the loss function, the mean squared errors from the true parameters are plotted.}
    \label{fig:learning_curve}
\end{figure}

Additionally, we compared the progression of stochastic optimization in Figure \ref{fig:learning_curve}. Each objective was optimized 100 times, and the means and the standard errors (68\% confidence interval) are represented by the lines and error bands. Again, we can see that the MLMC method and the randomized MLMC method converge to the smallest MSEs than others. Though the SUMO did not converge as fast as other estimators in this experiment, it converged to good solutions after sufficient time was elapsed, as shown in Table \ref{tab:accuracy}.

\subsection{Computational Efficiency of MLMC}

\begin{figure}
    \centering
    \includegraphics[width=\linewidth, trim=0mm 5mm 0mm 10mm, clip]{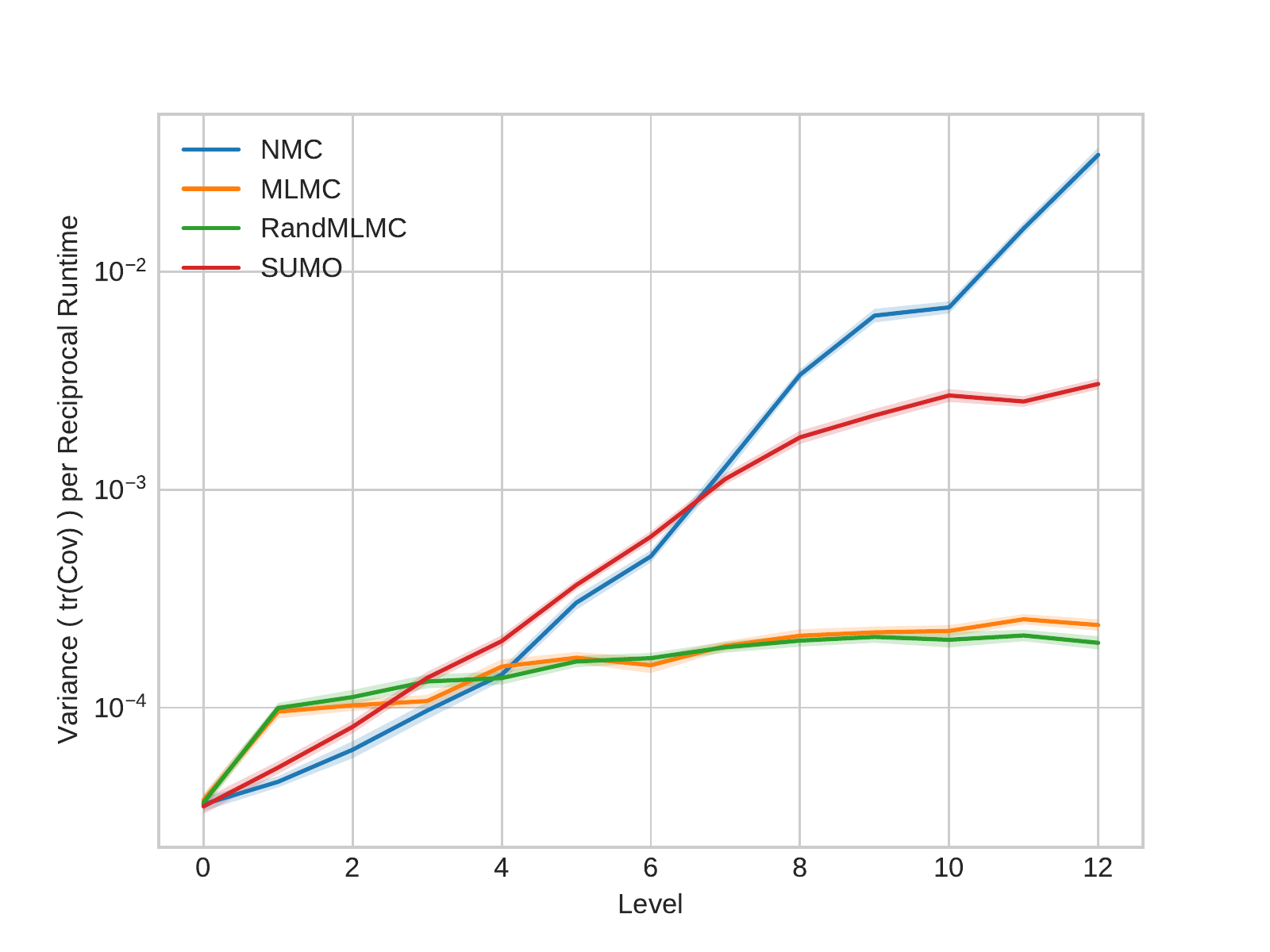}
    \caption{Computational efficiency of the gradient at each level. Two to the power of the level corresponds to the number of inner Monte Carlo samples, i.e. $2^L=K$.}
    \label{fig:efficiency}
\end{figure}

To quantitatively compare the computational efficiency of each estimator, we plotted the variance of the gradient estimator of $\nabla_\theta \Lcal_{2^L}$ against each level $L$, for a given computational cost (runtime) in Figure \ref{fig:efficiency}. As the variance of Monte Carlo estimators decreases reciprocally to the number of samples (or, the computational cost), we multiplied the variance of each estimator by the runtime the estimator spent to obtain a measure of computational efficiency. The Jackknife estimator is not compared, because its bias is not equal to those of the other estimators. In this experiment, we used the largest batch size that fits in the memory to ignore the implementational inefficiencies of our Python code. Unlike nested Monte Carlo estimator, the iteration over multiple levels in MLMC and SUMO cannot be written with basic array operations, and this causes runtime overhead when the batch size is small.

Theoretically, the variance per computational cost becomes $O(1)$ for the MLMC-based estimators, while $O(2^L)$ and $O(L^2)$ costs are required for the nested Monte Carlo and the SUMO, respectively. For large levels, the MLMC-based estimators are 10 to 100 times more efficient than other estimators. However, in the low-level regions ($L \leq 3$), although the corresponding objective is quite biased, the nested Monte Carlo estimator is the most efficient.

\section{CONCLUSIONS}

This paper introduced a new estimator for the model evidence and its gradient based on the MLMC sampling technique. In the theoretical analysis, we showed that the computational complexity of our MLMC estimator is an order of magnitude smaller than the standard nested Monte Carlo estimator and the estimator can be made unbiased while having finite variance and expected computational cost. This property is unprecedented by any other existing debiasing methods for the model evidence estimation. In the experiments, we confirmed that our MLMC estimator performs as expected from the theory and observed its superiority over the existing estimators. 

To our knowledge, the MLMC has almost never been applied in the field of machine learning. Thus, we believe that the exploration of other possible applications of MLMC to this field is very promising.

\begin{acknowledgements} 
K.I. acknowledges support by the Heiwa Nakajima Foundation and Physical Sciences Research Council (EPSRC) grant.
The work of T.G. is supported by JSPS KAKENHI Grant Number~20K03744.
\end{acknowledgements}

\bibliography{references}

\begin{thebibliography}{27}
\providecommand{\natexlab}[1]{#1}
\providecommand{\url}[1]{\texttt{#1}}
\expandafter\ifx\csname urlstyle\endcsname\relax
  \providecommand{\doi}[1]{doi: #1}\else
  \providecommand{\doi}{doi: \begingroup \urlstyle{rm}\Url}\fi

\bibitem[Anderson and Higham(2012)]{anderson2012multilevel}
David~F Anderson and Desmond~J Higham.
\newblock Multilevel {M}onte {C}arlo for continuous time {M}arkov chains, with
  applications in biochemical kinetics.
\newblock \emph{Multiscale Modeling \& Simulation}, 10\penalty0 (1):\penalty0
  146--179, 2012.

\bibitem[Bishop(2006)]{bishop2006pattern}
Christopher~M Bishop.
\newblock \emph{Pattern recognition and machine learning}.
\newblock springer, 2006.

\bibitem[Bujok et~al.(2015)Bujok, Hambly, and Reisinger]{bujok2015multilevel}
Karolina Bujok, BM~Hambly, and Christoph Reisinger.
\newblock Multilevel simulation of functionals of {B}ernoulli random variables
  with application to basket credit derivatives.
\newblock \emph{Methodology and Computing in Applied Probability}, 17\penalty0
  (3):\penalty0 579--604, 2015.

\bibitem[Burda et~al.(2015)Burda, Grosse, and
  Salakhutdinov]{burda2015importance}
Yuri Burda, Roger Grosse, and Ruslan Salakhutdinov.
\newblock Importance weighted autoencoders.
\newblock \emph{arXiv preprint arXiv:1509.00519}, 2015.

\bibitem[Cliffe et~al.(2011)Cliffe, Giles, Scheichl, and
  Teckentrup]{cliffe2011multilevel}
K~Andrew Cliffe, Mike~B Giles, Robert Scheichl, and Aretha~L Teckentrup.
\newblock Multilevel {M}onte {C}arlo methods and applications to elliptic
  {P}{D}{E}s with random coefficients.
\newblock \emph{Computing and Visualization in Science}, 14\penalty0
  (1):\penalty0 3, 2011.

\bibitem[Dempster et~al.(1977)Dempster, Laird, and Rubin]{dempster1977maximum}
Arthur~P Dempster, Nan~M Laird, and Donald~B Rubin.
\newblock Maximum likelihood from incomplete data via the em algorithm.
\newblock \emph{Journal of the Royal Statistical Society: Series B
  (Methodological)}, 39\penalty0 (1):\penalty0 1--22, 1977.

\bibitem[Dodwell et~al.(2015)Dodwell, Ketelsen, Scheichl, and
  Teckentrup]{dodwell2015hierarchical}
Tim~J Dodwell, Christian Ketelsen, Robert Scheichl, and Aretha~L Teckentrup.
\newblock A hierarchical multilevel {M}arkov {C}hain {M}onte {C}arlo algorithm
  with applications to uncertainty quantification in subsurface flow.
\newblock \emph{SIAM/ASA Journal on Uncertainty Quantification}, 3\penalty0
  (1):\penalty0 1075--1108, 2015.

\bibitem[Dua and Graff(2017)]{Dua:2019}
Dheeru Dua and Casey Graff.
\newblock {UCI} machine learning repository, 2017.
\newblock URL \url{http://archive.ics.uci.edu/ml}.

\bibitem[Giles(2008)]{giles2008multilevel}
Michael~B Giles.
\newblock Multilevel {M}onte {C}arlo path simulation.
\newblock \emph{Operations Research}, 56\penalty0 (3):\penalty0 607--617, 2008.

\bibitem[Giles(2015)]{giles2015multilevel}
Michael~B Giles.
\newblock Multilevel {M}onte {C}arlo methods.
\newblock \emph{Acta Numerica}, 24:\penalty0 259--328, 2015.

\bibitem[Giles and Goda(2019)]{giles2020decision}
Michael~B Giles and Takashi Goda.
\newblock Decision-making under uncertainty: using {M}{L}{M}{C} for efficient
  estimation of {E}{V}{P}{P}{I}.
\newblock \emph{Statistics and Computing}, 29\penalty0 (4):\penalty0 739--751,
  2019.

\bibitem[Goda et~al.(2020)Goda, Hironaka, and Iwamoto]{goda2020multilevel}
Takashi Goda, Tomohiko Hironaka, and Takeru Iwamoto.
\newblock Multilevel {M}onte {C}arlo estimation of expected information gains.
\newblock \emph{Stochastic Analysis and Applications}, 38\penalty0
  (4):\penalty0 581--600, 2020.

\bibitem[Goda et~al.(2021)Goda, Hironaka, Kitade, and Foster]{goda2021design}
Takashi Goda, Tomohiko Hironaka, Wataru Kitade, and Adam Foster.
\newblock Unbiased {M}{L}{M}{C} stochastic gradient-based optimization of
  {B}ayesian experimental designs.
\newblock \emph{arXiv preprint arXiv:2005.08414v2}, 2021.

\bibitem[Heinrich(1998)]{heinrich1998monte}
Stefan Heinrich.
\newblock Monte {C}arlo complexity of global solution of integral equations.
\newblock \emph{Journal of Complexity}, 14\penalty0 (2):\penalty0 151--175,
  1998.

\bibitem[Hensman et~al.(2013)Hensman, Fusi, and Lawrence]{hensman2013gaussian}
James Hensman, Nicolo Fusi, and Neil~D Lawrence.
\newblock Gaussian processes for big data.
\newblock \emph{arXiv preprint arXiv:1309.6835}, 2013.

\bibitem[Hironaka et~al.(2020)Hironaka, Giles, Goda, and
  Thom]{hironaka2020multilevel}
Tomohiko Hironaka, Michael~B Giles, Takashi Goda, and Howard Thom.
\newblock Multilevel {M}onte {C}arlo estimation of the expected value of sample
  information.
\newblock \emph{SIAM/ASA Journal on Uncertainty Quantification}, 8\penalty0
  (3):\penalty0 1236--1259, 2020.

\bibitem[Hoffman et~al.(2013)Hoffman, Blei, Wang, and
  Paisley]{hoffman2013stochastic}
Matthew~D Hoffman, David~M Blei, Chong Wang, and John Paisley.
\newblock Stochastic variational inference.
\newblock \emph{Journal of Machine Learning Research}, 14\penalty0
  (1):\penalty0 1303--1347, 2013.

\bibitem[Jordan et~al.(1999)Jordan, Ghahramani, Jaakkola, and
  Saul]{jordan1999introduction}
Michael~I Jordan, Zoubin Ghahramani, Tommi~S Jaakkola, and Lawrence~K Saul.
\newblock An introduction to variational methods for graphical models.
\newblock \emph{Machine learning}, 37\penalty0 (2):\penalty0 183--233, 1999.

\bibitem[Kingma and Ba(2014)]{kingma2014adam}
Diederik~P Kingma and Jimmy Ba.
\newblock Adam: A method for stochastic optimization.
\newblock \emph{arXiv preprint arXiv:1412.6980}, 2014.

\bibitem[Kingma and Welling(2013)]{kingma2013auto}
Diederik~P Kingma and Max Welling.
\newblock Auto-encoding variational {B}ayes.
\newblock \emph{arXiv preprint arXiv:1312.6114}, 2013.

\bibitem[Luo et~al.(2019)Luo, Beatson, Norouzi, Zhu, Duvenaud, Adams, and
  Chen]{luo2019sumo}
Yucen Luo, Alex Beatson, Mohammad Norouzi, Jun Zhu, David Duvenaud, Ryan~P
  Adams, and Ricky~TQ Chen.
\newblock {S}{U}{M}{O}: {U}nbiased estimation of log marginal probability for
  latent variable models.
\newblock 2019.

\bibitem[Nowozin(2018)]{nowozin2018debiasing}
Sebastian Nowozin.
\newblock Debiasing evidence approximations: {O}n importance-weighted
  autoencoders and {J}ackknife variational inference.
\newblock 2018.

\bibitem[Rhee and Glynn(2015)]{rhee2015unbiased}
Chang-han Rhee and Peter~W Glynn.
\newblock Unbiased estimation with square root convergence for {S}{D}{E}
  models.
\newblock \emph{Operations Research}, 63\penalty0 (5):\penalty0 1026--1043,
  2015.

\bibitem[Robbins and Monro(1951)]{robbins1951stochastic}
Herbert Robbins and Sutton Monro.
\newblock A stochastic approximation method.
\newblock \emph{The annals of mathematical statistics}, pages 400--407, 1951.

\bibitem[Robert and Casella(2013)]{robert2013monte}
Christian Robert and George Casella.
\newblock \emph{Monte {C}arlo statistical methods}.
\newblock Springer Science \& Business Media, 2013.

\bibitem[Snelson and Ghahramani(2006)]{snelson2006sparse}
Edward Snelson and Zoubin Ghahramani.
\newblock Sparse gaussian processes using pseudo-inputs.
\newblock In \emph{Advances in neural information processing systems}, pages
  1257--1264, 2006.

\bibitem[Wei and Tanner(1990)]{wei1990monte}
Greg~CG Wei and Martin~A Tanner.
\newblock A {M}onte {C}arlo implementation of the {E}{M} algorithm and the poor
  man's data augmentation algorithms.
\newblock \emph{Journal of the American statistical Association}, 85\penalty0
  (411):\penalty0 699--704, 1990.

\end{thebibliography}

\newpage
\onecolumn
\appendix
\allowdisplaybreaks

{\Large \textbf{Appendix: Efficient Debiased Evidence Estimation by Multilevel Monte Carlo Sampling}}

\section{A Review of Multilevel Monte Carlo methods}\label{app:review_MLMC}


\subsection{Background on MLMC methods}

MLMC has been originally introduced by \citet{heinrich1998monte} for parametric integration and then applied to the context of stochastic differential equations by \citet{giles2008multilevel}. Thereafter MLMC has been applied to a variety of subjects, partial differential equations with random fields \citep{cliffe2011multilevel}, continuous time Markov chains \citep{anderson2012multilevel}, Markov chain Monte Carlo sampling \citep{dodwell2015hierarchical}, and nested expectations. Applications of MLMC to nested expectations can be found in \citet{bujok2015multilevel}, \citet{giles2020decision}, \citet{hironaka2020multilevel} and \citet{goda2021design} among many others. As far as the authors know, no study has been conducted for applications to evidence estimation.

Roughly speaking, MLMC can be regarded as an extension of control variate technique. Let us consider a problem of estimating the expectation $\EE[f]$. The control variate technique is to find a good auxiliary random variable $g$ with known $\EE[g]$ and to estimate $\EE[f]$ by
\[ \EE[f]= \EE[f-g]+\EE[g]\approx \frac{1}{N}\sum_{k=1}^{N}(f-g)^{(k)}+\EE[g]. \]
When $\VV[f-g]\ll \VV[f]$ and the cost for computing $f-g$ is almost of the same order with that for $f$, one can expect a significant cost saving as compared to the standard Monte Carlo estimator $(1/N)\sum_{k=1}^{N}f^{(k)}$ to achieve the same mean squared accuracy. 

A two-level version of MLMC considers an auxiliary variable $g$ with not-necessarily known $\EE[g]$ and estimates $\EE[f]$ by
\[ \EE[f]= \EE[f-g]+\EE[g]\approx \frac{1}{N_1}\sum_{k=1}^{N_1}(f-g)^{(k)}+\frac{1}{N_2}\sum_{k=1}^{N_2}g^{(k)}. \]
Here, $\EE[f-g]$ and $\EE[g]$ are estimated independently with different numbers of samples. Now if we have $g \approx f$ such that $\VV[g]\approx \VV[f]\gg \VV[f-g]$ and the cost for computing $g$, denoted by $c_g$ is much smaller than that for $f$, denoted by $c_f$, the two-level Monte Carlo estimator compares favorably with the standard one. In fact, the variance and the cost of the two-level Monte Carlo estimator are given by
\[ \frac{\VV[f-g]}{N_1}+\frac{\VV[g]}{N_2}\quad \textrm{and}\quad (c_f+c_g)N_1+c_gN_2, \]
respectively, whereas those of the standard one are 
\[ \frac{\VV[f]}{N} \quad \textrm{and}\quad c_fN. \]
The method of Lagrange multipliers leads to optimal numbers of samples $N_1$ and $N_2$ which minimize the cost of the two-level Monte Carlo estimator with its variance equal to $\VV[f]/N$:
\begin{align*}
    N_1 & = N\frac{\sqrt{(c_f+c_g)\VV[f-g]}+\sqrt{c_g\VV[g]}}{\VV[f]}\sqrt{\frac{\VV[f-g]}{c_f+c_g}},\\
    N_2 & = N\frac{\sqrt{(c_f+c_g)\VV[f-g]}+\sqrt{c_g\VV[g]}}{\VV[f]}\sqrt{\frac{\VV[g]}{c_g}},
\end{align*} 
respectively, giving the total cost
\[ (c_f+c_g)N_1+c_gN_2 = N\frac{\left(\sqrt{(c_f+c_g)\VV[f-g]}+\sqrt{c_g\VV[g]}\right)^2}{\VV[f]}. \]
Recall that we are in the situation where $c_g\ll c_f$ and $\VV[g]\approx \VV[f]\gg \VV[f-g]$.
If $(c_f+c_g)\VV[f-g]> c_g\VV[g]$, the total cost is approximately equal to $c_f N\left(\VV[f-g]/\VV[f]\right)$. On the other hand, if $(c_f+c_g)\VV[f-g]< c_g\VV[g]$, the total cost is approximately equal to $c_g N$. In both cases, the cost for the two-level Monte Carlo estimator is smaller than that for the standard one, which is $c_f N$.

The MLMC estimator is a natural extension of this two-level Monte Carlo estimator. Let us consider a sequence of auxiliary random variables $f_0,f_1,\ldots$. Then the following telescoping sum holds:
\[\EE[f_L] = \EE[f_0]+\EE[f_1-f_0]+\cdots + \EE[f_L-f_{L-1}]= \sum_{\ell=0}^{L}\EE[f_{\ell}-f_{\ell-1}], \]
where we write $f_{-1}\equiv 0$. By estimating each of the terms on the rightmost side above independently, the MLMC estimator is given by
\[ \sum_{\ell=0}^{L}\frac{1}{N_{\ell}}\sum_{k=1}^{N_{\ell}}(f_{\ell}-f_{\ell-1})^{(k)},\]
with different numbers of samples $N_0,N_1,\ldots,N_L$. More generally, if we have a sequence of "correction" random variables $\Delta f_0,\Delta f_1,\ldots$ such that $\EE[\Delta f_0]=\EE[f_0]$ and $\EE[\Delta f_{\ell}]=\EE[f_{\ell}-f_{\ell-1}]$ for $\ell\geq 1$, the MLMC estimator reduces to
\[ \sum_{\ell=0}^{L}\frac{1}{N_{\ell}}\sum_{k=1}^{N_{\ell}}(\Delta f_{\ell})^{(k)}.\]
The key quantities in the MLMC estimator are
\begin{enumerate}
    \item the decay of the bias $\EE[f_{\ell}-f_{\infty}]$, 
    \item the decay of $\VV[\Delta f_{\ell}]$, and
    \item the expected cost per one evaluation of $\Delta f_{\ell}$, denoted by $C_{\ell}$.
\end{enumerate}
The first item determines the maximum level $L$ of the MLMC estimator such that the bias is sufficiently small. The second and third items determine the optimal allocation of the numbers of samples $N_0,N_1,\ldots,N_L$ such that the total cost is minimized while achieving a given variance. Here, the variance and the total cost of the MLMC estimator are given by
\[ \sum_{\ell=0}^{L}\frac{\VV[\Delta f_{\ell}]}{N_{\ell}}\quad \textrm{and}\quad \sum_{\ell=0}^{L}C_{\ell}N_{\ell},\]
respectively.

\subsection{Basic Theory of MLMC methods}
Let us assume that the following conditions hold for the key quantities of the MLMC estimator. That is, assume that there exist positive constants $\alpha,\beta,\gamma,c_1,c_2,c_3$ such that
\begin{enumerate}
    \item $\alpha\geq \min(\beta,\gamma)/2$, 
    \item $|\EE[f_{\ell}-f_{\infty}]|\leq c_12^{-\alpha \ell}$,
    \item $\VV[\Delta f_{\ell}]\leq c_2 2^{-\beta \ell}$, and
    \item $C_{\ell}\leq c_3 2^{\gamma \ell}$.
\end{enumerate}
Then we have the following basic MLMC theorem, proven by \citet{giles2008multilevel} for the case $\gamma=1$ and by \citet{cliffe2011multilevel} for a general $\gamma$. Here, we recall that our Theorem~\ref{thm:mlmc} corresponds to the case $\gamma=1$ as we use $2^{\ell}$ inner Monte Carlo samples to define $\Delta \Lcalhat_{1,2^{\ell}}$. 
\begin{theorem}\label{thm:mlmc_general}
If the above conditions hold, for any given accuracy $\varepsilon<\exp(-1)$, there exists a positive constant $c_4$ such that there are the corresponding maximum level $L$ and the numbers of samples $N_0,N_1,\ldots, N_L$ for which the mean squared error of the MLMC estimator for $\EE[f_{\infty}]$ is less than $\varepsilon^2$ with the total cost $C$ bounded by
\[ \EE[C]\leq \begin{cases} c_4\varepsilon^{-2}, & \beta>\gamma, \\ c_4\varepsilon^{-2}|\log \varepsilon^{-1}|^2, & \beta=\gamma, \\ c_4\varepsilon^{-2-(\gamma-\beta)/\alpha}, & \beta<\gamma. \end{cases} \]
\end{theorem}

In what follows, we give a rough sketch on the proof of Theorem~\ref{thm:mlmc_general}.

First we note that the mean squared error of the MLMC estimator is decomposed into the squared bias and the variance:
\begin{align}\label{eq:variance-bias}
 \left| \EE[f_{L}-f_{\infty}]\right|^2+\sum_{\ell=0}^{L}\frac{\VV[\Delta f_{\ell}]}{N_{\ell}}.  
\end{align}
Thus, in order to make \eqref{eq:variance-bias} less than or equal to $\varepsilon^2$, it suffices to have
\[ \left| \EE[f_{L}-f_{\infty}]\right|\leq  \frac{\varepsilon}{\sqrt{2}}\quad \textrm{and}\quad \sum_{\ell=0}^{L}\frac{\VV[\Delta f_{\ell}]}{N_{\ell}}\leq \frac{\varepsilon^2}{2},\]
simultaneously. Given the second condition on the decay of the bias, the first inequality holds for $L$ satisfying
\[ c_12^{-\alpha L}\leq \frac{\varepsilon}{\sqrt{2}}.\]
Therefore we choose
\begin{align}\label{eq:maximum_level} L= \left\lceil \frac{\log_2(\sqrt{2}c_1\varepsilon^{-1})}{\alpha} \right\rceil .\end{align}
An argument similar to that used in the two-level Monte Carlo estimator leads to an optimal allocation of the numbers of samples $N_0,N_1,\ldots,N_L$ which minimize the cost of the MLMC estimator with its variance no larger than $\varepsilon^2/2$:
\[ N_{\ell} = \left\lceil 2\varepsilon^{-2} \sqrt{\frac{\VV[\Delta f_{\ell}]}{C_{\ell}}}\sum_{\ell'=0}^{L}\sqrt{C_{\ell'} \VV[\Delta f_{\ell'}]}\right\rceil. \]
The corresponding total cost is bounded above by
\begin{align*}
    \sum_{\ell=0}^{L}C_{\ell}N_{\ell} & \leq \sum_{\ell=0}^{L}C_{\ell}\left( 1+2\varepsilon^2 \sqrt{\frac{\VV[\Delta f_{\ell}]}{C_{\ell}}}\sum_{\ell'=0}^{L}\sqrt{C_{\ell'} \VV[\Delta f_{\ell'}]}\right) \\
    & = \sum_{\ell=0}^{L}C_{\ell} + 2\varepsilon^{-2}\left( \sum_{\ell=0}^{L}\sqrt{C_{\ell} \VV[\Delta f_{\ell}]}\right)^2 \\
    & \leq c_3\sum_{\ell=0}^{L}2^{\gamma \ell} + 2c_2c_3\varepsilon^{-2}\left( \sum_{\ell=0}^{L}2^{-(\beta-\gamma)\ell/2}\right)^2.
\end{align*}
The remaining task to complete the proof is to substitute the choice of the maximum level \eqref{eq:maximum_level} into the rightmost side above and give bounds for the respective cases $\beta>\gamma$, $\beta=\gamma$, and $\beta<\gamma$ through an elementary computation.

\section{Comparison of NMC, SUMO and MLMC}\label{app:comparison}
To illustrate the difference in efficiency between the several estimation methods of the model evidence, we summarize their convergence properties in Table \ref{tab:summary_ests}. Here, we consider the case where MLMC is efficiently applicable ($\alpha=1$ and $\beta>1$). The results for the nested Monte Carlo estimator and the MLMC estimator are established in the main article. The variance of the Jackknife estimator, called JVI, is hard to analyze in general as discussed in \citet{nowozin2018debiasing}. For the SUMO estimator, we derive the variance and cost in the following proposition, as they are not provided in its original article by \citet{luo2019sumo}. Our analysis on the SUMO estimator assumes an estimator with hard truncation where the random level $\mathcal{K}$ is upper bounded. This is because the cost and the variance of the SUMO estimator with soft truncation become both infinities. Though our analysis does not consider the gradient of the SUMO, a similar analysis reveals that the same efficiency holds for the gradient counterpart. 
\begin{table}[ht]
    \caption{Bias, variance, cost and computational efficiency of gradient estimator of model evidence using different estimation methods, in the case of $\alpha=1$ and $\beta>1$.}
    \centering
    \label{tab:summary_ests}
    \begin{tabular}{ccccc}
        \toprule
        & NMC & MLMC & SUMO & JVI (order $m$) \\
        \midrule
        Bias & $\Ocal(1/K)$ & $\Ocal(2^{-L})$ & $\Ocal(1/K)$ & $\Ocal(1/K^{m+1})$ \\
        Variance per data point & $\Ocal(1)$ & $\Ocal(1)$ & $\Ocal(\log K)$ & hard to know \\
        Expected Cost per data point & $\Ocal(K)$ & $\Ocal(1)$ & $\Ocal(\log K)$ & $\Ocal(K)$ \\
        Efficiency (Cost $\times$ Var) & $\Ocal(K)$ & $\Ocal(1)$ & $\Ocal((\log K)^2)$ & hard to know \\
        \bottomrule
    \end{tabular}
    \label{tab:my_label}
\end{table}

\begin{proposition}
Consider the SUMO estimator with hard truncation at random level $\mathcal{K} \leq K$, 
\[\Lcalhat^{\text{SUMO}}_{1, K}(X) = \sum_{k=1}^K\frac{\mathbbm{1}_{\{k \leq \mathcal{K}\}}}{\PP(k \leq \tilde{\mathcal{K}})}\Delta_k,\] 
where we let $\Delta_k := \widehat{[\Lcal_k - \Lcal_{k-1}]}(X)= \Lcalhat_K(X; Z_{1:K}) - \Lcalhat_{K-1}(X; Z_{1:(K-1)})$ for notational simplicity. Assume that the parameter $\alpha$ in Theorem \ref{thm:mlmc} satisfies $\alpha=1$ and that $\mathcal{K}$ is bounded above by $K$ and follows $\mathbb{P}(k\leq\mathcal{K})=\frac{1}{k}$. $\tilde{\mathcal{K}}$ is distributed identically to $\mathcal{K}$.
Then, the variance and cost of the estimator $\Lcalhat^{\text{SUMO}}_{1, K}$ both become $\Ocal(\log K)$.
\end{proposition}

\begin{proof}
As $\mathbb{P}(k\leq\mathcal{K})=\frac{1}{k}$, we have $\mathbb{P}(k=\mathcal{K}<K)=\mathbb{P}(k\leq\mathcal{K}<K)-\mathbb{P}(k+1\leq \mathcal{K}<K)=\frac{1}{k(k+1)}$. From this, it is easy to calculate the expected cost $\EE[C_{\Lcalhat^{\text{SUMO}}_{1, K}}]$ as  
\begin{align*}
    \EE[C_{\Lcalhat^{\text{SUMO}}_{1, K}}] 
    = \EE_{\mathcal{K}}[\EE[C_{\Lcalhat^{\text{SUMO}}_{1, K}}|\mathcal{K}]] = \EE_{\mathcal{K}}[\Ocal(\mathcal{K})] = \sum_{k=1}^K \frac{1}{k(k+1)}\Ocal(k) = \Ocal(\log K).
\end{align*}
Though the cost of computing each $\Delta_k$ is $\Ocal(k)$, the cost of computing $\Lcalhat^{\text{SUMO}}_{1, K}(X) = \sum_{k=1}^K\frac{\mathbbm{1}_{\{k \leq \mathcal{K}\}}}{\PP(k \leq \tilde{\mathcal{K}})}\Delta_k$ given $\mathcal K$ becomes $\Ocal(\mathcal{K})$ by using an efficient algorithm discussed in \citet{luo2019sumo}, from which the second equality follows.

Now, we show the convergence rate of the variance of $\Lcalhat^{\text{SUMO}}_{1, K}$. When $\alpha=1$, \citet{luo2019sumo} showed that the convergence behavior of $\Delta_k$ becomes: 
\[ \EE[\Delta_k \Delta_l] = \begin{cases}
\Ocal(k^{-2}) \quad \text{if } k=l\\
\Ocal(k^{-2} l^{-2}) \quad \text{otherwise}.
\end{cases}\]
Also, as we show later, we have 
\begin{equation}
    \EE\Delta_k = \Ocal(k^{-2})\label{eq:EDelta_k}.
\end{equation}

Using these properties, we can analyze the order of the variance of $\Lcalhat^{\text{SUMO}}_{1, K}$ as:
\begin{align*}
\mathrm{Var}\left[\Lcalhat^{\text{SUMO}}_{1, K}\right]
&= \mathrm{Var}_{\mathcal{K}, \Delta_{1:K}} 
\left[ \sum_{k=1}^{K} \frac { \mathbbm{1}_{(k\leq\mathcal{K})} } 
{ \mathbb{P}(k\leq\tilde{\mathcal{K}}) } \Delta_k \right] \\
&= \EE_{\mathcal{K}}\mathrm{Var}_{\Delta_{1:K}}
\left[ \sum_{k=1}^{\mathcal{K}} k \Delta_k \bigg| \mathcal {K}\right]
 + \mathrm{Var}_{\mathcal{K}}\EE_{\Delta_{1:K}}
\left[ \sum_{k=1}^{\mathcal{K}} k \Delta_k \bigg| \mathcal {K}\right] \\
&= \EE_{\mathcal{K}}\left[\sum_{k=1}^{\mathcal{K}} \sum_{l=1}^{\mathcal{K}} kl \cdot \mathrm{Cov}_{\Delta_{1:K}}
\left[ \Delta_k\Delta_l | \mathcal {K}\right] \right]
 + \mathrm{Var}_{\mathcal{K}} \left[\sum_{k=1}^{\mathcal{K}} k \EE_{\Delta_{1:K}}
\left[ \Delta_k | \mathcal {K}\right] \right]\\
&\leq \EE_{\mathcal{K}}\left[\sum_{k=1}^{\mathcal{K}} \sum_{l=1}^{\mathcal{K}} kl 
\cdot \left( \EE\Delta_k\Delta_l - \EE\Delta_k\EE\Delta_l\right) \right]
 + \mathrm{Var}_{\mathcal{K}} \left[\sum_{k=1}^{\mathcal{K}} k \EE \Delta_k \right]\\
&= \EE_{\mathcal{K}}\left[\sum_{\substack{k,l=1\\k\neq l}}^{\mathcal{K}} kl 
\cdot \Ocal(k^{-2}l^{-2}) + \sum_{k=1}^\mathcal{K} k^2\Ocal(k^{-2})\right]
 + \mathrm{Var}_{\mathcal{K}} \left[\sum_{k=1}^{\mathcal{K}} k \Ocal(k^{-2}) \right]\\
&= \EE_{\mathcal{K}} \left[\Ocal\left((\log \mathcal{K})^2 \right) + \Ocal(\mathcal{K})\right]
 + \mathrm{Var}_{\mathcal{K}} \left[ \Ocal(\log \mathcal{K}) \right]\\
&= \Ocal(\log K)
\end{align*}

Now we show the property by following the proof techniques used in \citet{nowozin2018debiasing} and \citet{luo2019sumo}. Let $w_i= \frac{p(x|Z_i)p(Z_i)}{q(Z_i|x)}$ and $Y_k=\frac{1}{k}\sum_{i=1}^k w_i$ so that $\EE Y_k=\EE w  = \mu$. Let us consider the Taylor expansion:
\[\log Y_k = \log \left[\mu + (Y_k - \mu)\right]=\log \mu - \sum_{t=1}^\infty\frac{(-1)^t}{t\mu^t}(Y_k-\mu)^t.\]
Here, the condition $|Y_k-\mu| < \mu$ is assumed for the convergence of power series. Now, we can calculate the order of $\EE\Delta_k$ using the third-order expansion as 
\begin{align*}
   \EE \Delta_k 
   & = \EE\left[\log \mu - \sum_{t=1}^\infty\frac{(-1)^t}{t\mu^t}(Y_k-\mu)^t - \log \mu + \sum_{t=1}^\infty\frac{(-1)^t}{t\mu^t}(Y_{k-1}-\mu)^t\right] \\
   & = \EE\left[ \frac{1}{\mu}(Y_k - \mu) - \frac{1}{\mu}(Y_{k-1} - \mu) - \frac{1}{2\mu^2}(Y_k - \mu)^2 + \frac{1}{2\mu^2}(Y_{k-1} - \mu)^2 + \frac{1}{6\mu^3}(Y_{k} - \mu)^3 - \frac{1}{6\mu^3}(Y_{k-1} - \mu)^3\right]   + \Ocal(k^{-2})\\
   &=  0 - 0 - \frac{1}{2k\mu^2}\VV[w] + \frac{1}{2(k-1)\mu^2}\VV[w]  + \frac{1}{6k^2\mu^3}\EE(w - \mu)^3 - \frac{1}{6(k-1)^2\mu^3}\EE(w - \mu)^3  + \Ocal(k^{-2})\\
   &= \frac{1}{2k(k-1)\mu^2}\VV[w] - \frac{2k-1}{6k^2(k-1)^2\mu^3}\EE(w - \mu)^3  + \Ocal(k^{-2})\\
   &= \Ocal(k^{-2}),
\end{align*}
where we used the results on the second and third central moments for the sample mean as stated in \citet{nowozin2018debiasing}.
\end{proof}

Though the property \eqref{eq:EDelta_k} is somewhat counter-intuitive compared to $\EE\Delta_k^2=\Ocal(k^{-2})$, it makes sense by considering the decay of the bias, $|\Lcal_{K}-\Lcal(x_{1:N})|\leq c_1K^{-\alpha}$ with $\alpha=1$. By the convexity of the log function, Jensen's inequality leads to $\EE\Delta_k \geq 0$ for any $k$, and the bias can be written as
\[\Lcal_{K}-\Lcal(x_{1:N}) = - \sum_{k=K+1}^\infty \EE\Delta_k = \sum_{k=K+1}^\infty \Ocal(k^{-2}) = \Ocal(K^{-1}).\]
Thus, the order of the bias agrees with the condition $\alpha=1$.

\section{Proofs of Theorems 2 and 3}\label{app:proofs}

Here, we give proofs of Theorems~\ref{thm:evidence} and \ref{thm:gradient}. We note that the essential argument for the former theorem follows from \cite{goda2020multilevel}, whereas the latter follows from \cite{hironaka2020multilevel} and \cite{goda2021design}.

\begin{proof}[Proof of Theorem~\ref{thm:evidence}]
Recalling that $\Lcalhat^{(a)}_{2^{\ell-1}}$ and $\Lcalhat^{(b)}_{2^{\ell-1}}$ are computed by using the first and second halves of the samples $Z_{1},\ldots,Z_{2^{\ell}}$ to compute $\Lcalhat_{2^{\ell}}$, respectively, we have
\begin{align*}\label{eq:couple1} e^{\Lcalhat_{2^{\ell}}} = \frac{e^{\Lcalhat^{(a)}_{2^{\ell-1}}}+e^{\Lcalhat^{(b)}_{2^{\ell-1}}}}{2} = \frac{1}{2^\ell} \sum_{k=1}^{2^\ell}  \frac{p(X|Z_{ k})p(Z_{k})}{q(Z_{k}; X)}.
\end{align*}
This equality leads to
\begin{align*}
    \Delta\Lcalhat_{2^{\ell}} & = \Lcalhat_{2^{\ell}}- - \frac{\Lcalhat_{2^{\ell-1}}^{(a)}+\Lcalhat_{2^{\ell-1}}^{(b)}}{2} \\
    & = \Lcalhat_{2^{\ell}} -\Lcal-e^{\Lcalhat_{2^{\ell}}-\Lcal} - \frac{\Lcalhat^{(a)}_{2^{\ell-1}} -\Lcal-e^{\Lcalhat^{(a)}_{2^{\ell-1}}-\Lcal}}{2} -- \frac{\Lcalhat^{(b)}_{2^{\ell-1}} -\Lcal-e^{\Lcalhat^{(b)}_{2^{\ell-1}}-\Lcal}}{2} \\
    & = \log \left(e^{\Lcalhat_{2^{\ell}}-\Lcal}\right)-e^{\Lcalhat_{2^{\ell}}-\Lcal}+1 - \frac{\log \left( e^{\Lcalhat^{(a)}_{2^{\ell-1}}-\Lcal}\right)-e^{\Lcalhat^{(a)}_{2^{\ell-1}}-\Lcal}+1}{2} - \frac{\log \left( e^{\Lcalhat^{(b)}_{2^{\ell-1}}-\Lcal}\right)-e^{\Lcalhat^{(b)}_{2^{\ell-1}}-\Lcal}+1}{2}.
\end{align*}
Using Jensen's inequality and an elementary inequality 
\[ |\log x-x+1|\leq |x-1|^r \max(-\log x ,1),\]
which holds for $x>0$ and $1\leq r\leq 2$, we obtain a bound
\begin{align*}
    \left(\Delta\Lcalhat_{2^{\ell}}\right)^2 & \leq  2\left(\log \left(e^{\Lcalhat_{2^{\ell}}-\Lcal}\right)-e^{\Lcalhat_{2^{\ell}}-\Lcal}+1\right)^2 \\
    & \quad + \left(\log \left( e^{\Lcalhat^{(a)}_{2^{\ell-1}}-\Lcal}\right)-e^{\Lcalhat^{(a)}_{2^{\ell-1}}-\Lcal}+1\right)^2 + \left(\log \left( e^{\Lcalhat^{(b)}_{2^{\ell-1}}-\Lcal}\right)-e^{\Lcalhat^{(b)}_{2^{\ell-1}}-\Lcal}+1\right)^2 \\
    & \leq 2\left| e^{\Lcalhat_{2^{\ell}}-\Lcal}-1\right|^{2r}\max\left( -\log e^{\Lcalhat_{2^{\ell}}-\Lcal}, 1\right)^2 \\
    & \quad +\left| e^{\Lcalhat^{(a)}_{2^{\ell-1}}-\Lcal}-1\right|^{2r}\max\left( -\log e^{\Lcalhat^{(a)}_{2^{\ell-1}}-\Lcal}, 1\right)^2+\left| e^{\Lcalhat^{(b)}_{2^{\ell-1}}-\Lcal}-1\right|^{2r}\max\left( -\log e^{\Lcalhat^{(b)}_{2^{\ell-1}}-\Lcal}, 1\right)^2.
\end{align*}
Applying H\"{o}lder inequality with the exponents $t/(t-2)$ and $t/2$ and choosing $r=\min(\frac{s(t-2)}{2t}, 2)$, we have
\begin{align*}
    \VV\left[ \Delta\Lcalhat_{2^{\ell}}\right]  & \leq \EE\left[\left( \Delta\Lcalhat_{2^{\ell}}\right)^2\right] \\
    & \leq 2\left(\EE\left[\left| e^{\Lcalhat_{2^{\ell}}-\Lcal}-1\right|^{2rt/(t-2)}\right]\right)^{(t-2)/t}\left(\EE\left[\max\left( -\log e^{\Lcalhat_{2^{\ell}}-\Lcal}, 1\right)^{t}\right]\right)^{2/t} \\
    & \quad + \left(\EE\left[\left| e^{\Lcalhat^{(a)}_{2^{\ell-1}}-\Lcal}-1\right|^{2rt/(t-2)}\right]\right)^{(t-2)/t}\left(\EE\left[\max\left( -\log e^{\Lcalhat^{(a)}_{2^{\ell-1}}-\Lcal}, 1\right)^{t}\right]\right)^{2/t} \\
    & \quad + \left(\EE\left[\left| e^{\Lcalhat^{(b)}_{2^{\ell-1}}-\Lcal}-1\right|^{2rt/(t-2)}\right]\right)^{(t-2)/t}\left(\EE\left[\max\left( -\log e^{\Lcalhat^{(b)}_{2^{\ell-1}}-\Lcal}, 1\right)^{t}\right]\right)^{2/t}.
\end{align*}

Let us focus on the first term on the rightmost side above, as the other two terms can be shown bounded in the same manner. Using the Marcinkiewicz-Zygmund inequality, the first factor is bounded above by
\begin{align*}
    \EE\left[\left| e^{\Lcalhat_{2^{\ell}}-\Lcal}-1\right|^{2rt/(t-2)}\right] & = \EE\left[\left| \frac{1}{2^\ell} \sum_{k=1}^{2^\ell}  \frac{p(X|Z_{ k})p(Z_{k})}{p(X)q(Z_{k}; X)}-1\right|^{2rt/(t-2)}\right] \\
    & \leq \frac{C_{2rt/(t-2)}}{2^{\ell rt/(t-2)}}\EE_{X,Z}\left[\left| \frac{p(X|Z)p(Z)}{p(X)q(Z; X)}-1\right|^{2rt/(t-2)}\right],
\end{align*}
with some positive constant $C_{2rt/(t-2)}$. Here, the last expectation is assumed to be finite in the theorem, as the exponent $2rt/(t-2)$ is less than or equal to $s$ by our choice of $r$. Moreover, noting that the function $x\to \max(-\log x,1)$ is convex and applying Jensen's inequality twice, we see that the second factor is bounded above by
\begin{align*}
    \EE\left[\max\left( -\log e^{\Lcalhat_{2^{\ell}}-\Lcal}, 1\right)^{t}\right] & = \EE\left[\max\left( -\log \frac{1}{2^\ell} \sum_{k=1}^{2^\ell}  \frac{p(X|Z_{ k})p(Z_{k})}{p(X)q(Z_{k}; X)}, 1\right)^{t}\right] \\
    & \leq \EE\left[\left(\frac{1}{2^\ell} \sum_{k=1}^{2^\ell}\max\left( -\log  \frac{p(X|Z_{ k})p(Z_{k})}{p(X)q(Z_{k}; X)}, 1\right)\right)^{t}\right] \\
    & \leq \EE\left[\frac{1}{2^\ell} \sum_{k=1}^{2^\ell}\max\left( -\log   \frac{p(X|Z_{ k})p(Z_{k})}{p(X)q(Z_{k}; X)}, 1\right)^{t}\right] \\
    & \leq \EE\left[\frac{1}{2^\ell} \sum_{k=1}^{2^\ell}\left( \left|\log  \frac{p(X|Z_{ k})p(Z_{k})}{p(X)q(Z_{k}; X)}\right|^t +1\right)\right]  = \EE_{X,Z}\left[ \left|\log  \frac{p(X|Z)p(Z)}{p(X)q(Z; X)}\right|^t +1\right],
\end{align*}
where the last quantity is assumed finite in the theorem. Therefore, the variance of $\Delta\Lcalhat_{2^{\ell}}$ is shown to decay with the order of $2^{-r\ell}$ with $r=\min(\frac{s(t-2)}{2t}, 2)$. This directly implies that 
\[ \beta = \min\left(\frac{s(t-2)}{2t}, 2\right).\]
The result for the other parameter $\alpha$ can be shown similarly.
\end{proof}

\begin{proof}[Proof of Theorem~\ref{thm:gradient}]
Here, again, we only show a proof for the result on the parameter $\beta$, as the result for $\alpha$ can be shown similarly. Let us recall that the noisy gradient $\nabla_{\theta}\Lcalhat_{2^{\ell}}$ is given by the ratio
\[ \nabla_{\theta}\Lcalhat_{2^{\ell}}=\frac{\displaystyle \frac{1}{2^{\ell}}\sum_{k=1}^{2^{\ell}}\frac{ \nabla_{\theta}\left(p(X| Z_{k})p(Z_{k})\right)}{q(Z_{k};X)}}{\displaystyle \frac{1}{2^{\ell}}\sum_{k=1}^{2^{\ell}}\frac{p(X| Z_{k})p(Z_{k})}{q(Z_{k};X)}}.\]
In what follows, we write
\begin{align*}
    & \left[\nabla_{\theta}\Lcalhat_{2^{\ell}}\right]_N = \frac{1}{2^{\ell}}\sum_{k=1}^{2^{\ell}}\frac{ \nabla_{\theta}\left(p(X| Z_{k})p(Z_{k})\right)}{q(Z_{k};X)},\quad \left[\nabla_{\theta}\Lcalhat_{2^{\ell}}\right]_D = \frac{1}{2^{\ell}}\sum_{k=1}^{2^{\ell}}\frac{p(X| Z_{k})p(Z_{k})}{q(Z_{k};X)}, \\
    & \left[\nabla_{\theta}\Lcalhat^{(a)}_{2^{\ell-1}}\right]_N = \frac{1}{2^{\ell-1}}\sum_{k=1}^{2^{\ell-1}}\frac{ \nabla_{\theta}\left(p(X| Z_{k})p(Z_{k})\right)}{q(Z_{k};X)},\quad \left[\nabla_{\theta}\Lcalhat^{(a)}_{2^{\ell-1}}\right]_D = \frac{1}{2^{\ell-1}}\sum_{k=1}^{2^{\ell-1}}\frac{p(X| Z_{k})p(Z_{k})}{q(Z_{k};X)}, \\
    & \left[\nabla_{\theta}\Lcalhat^{(b)}_{2^{\ell-1}}\right]_N = \frac{1}{2^{\ell-1}}\sum_{k=2^{\ell-1}+1}^{2^{\ell}}\frac{ \nabla_{\theta}\left(p(X| Z_{k})p(Z_{k})\right)}{q(Z_{k};X)},\quad \left[\nabla_{\theta}\Lcalhat^{(b)}_{2^{\ell-1}}\right]_D = \frac{1}{2^{\ell-1}}\sum_{k=2^{\ell}+1}^{2^{\ell}}\frac{p(X| Z_{k})p(Z_{k})}{q(Z_{k};X)}.
\end{align*}
Here, we have
\[ \nabla_{\theta}\Delta \Lcalhat_{2^{\ell}} = \nabla_{\theta}\Lcalhat_{2^{\ell}}-\frac{\nabla_{\theta}\Lcalhat^{(a)}_{2^{\ell-1}}+\nabla_{\theta}\Lcalhat^{(b)}_{2^{\ell-1}}}{2}, \]
wherein, with the above notation, each term on the right-hand side is given by
\[ \nabla_{\theta}\Lcalhat_{2^{\ell}}=\frac{\left[\nabla_{\theta}\Lcalhat_{2^{\ell}}\right]_N}{\left[\nabla_{\theta}\Lcalhat_{2^{\ell}}\right]_D},\quad \nabla_{\theta}\Lcalhat^{(a)}_{2^{\ell-1}}=\frac{\left[\nabla_{\theta}\Lcalhat^{(a)}_{2^{\ell-1}}\right]_N}{\left[\nabla_{\theta}\Lcalhat^{(a)}_{2^{\ell-1}}\right]_D},\quad \textrm{and}\quad \nabla_{\theta}\Lcalhat^{(b)}_{2^{\ell-1}}=\frac{\left[\nabla_{\theta}\Lcalhat^{(b)}_{2^{\ell-1}}\right]_N}{\left[\nabla_{\theta}\Lcalhat^{(b)}_{2^{\ell-1}}\right]_D}.\]
It is obvious that the following equalities hold:
\begin{align}\label{eq:coupling2}
    \left[\nabla_{\theta}\Lcalhat_{2^{\ell}}\right]_N = \frac{\left[\nabla_{\theta}\Lcalhat^{(a)}_{2^{\ell-1}}\right]_N+\left[\nabla_{\theta}\Lcalhat^{(b)}_{2^{\ell-1}}\right]_N}{2},\quad \textrm{and} \quad \left[\nabla_{\theta}\Lcalhat_{2^{\ell}}\right]_D  = \frac{\left[\nabla_{\theta}\Lcalhat^{(a)}_{2^{\ell-1}}\right]_D+\left[\nabla_{\theta}\Lcalhat^{(b)}_{2^{\ell-1}}\right]_D}{2}.
\end{align} 

Now let us consider an event
\[ A := \left\{ \left|\frac{\left[\nabla_{\theta}\Lcalhat^{(a)}_{2^{\ell-1}}\right]_D}{p(X)}-1 \right|>\frac{1}{2}\right\}\, \bigcup\, \left\{ \left|\frac{\left[\nabla_{\theta}\Lcalhat^{(b)}_{2^{\ell-1}}\right]_D}{p(X)}-1 \right|>\frac{1}{2}\right\}. \]
Then we have
\begin{align}\label{eq:cases} \EE\left[ \|\nabla_{\theta}\Delta \Lcalhat_{2^{\ell}}\|^2_2\right] = \EE\left[ \|\nabla_{\theta}\Delta \Lcalhat_{2^{\ell}}\|^2_2 \bsone_A\right] + \EE\left[ \|\nabla_{\theta}\Delta \Lcalhat_{2^{\ell}}\|^2_2 \bsone_{A^c}\right],\end{align}
where $\bsone_{A}$ denotes the indicator function of the event $A$ and $A^c$ denotes the complement of $A$. Thus it suffices to prove that each of the two terms on the right-hand side of \eqref{eq:cases} decays with an order no slower than $2^{-\beta \ell}$ with $\beta=\min(s/2,2)$.
 
Consider the first term on the right-hand side of \eqref{eq:cases}. As we assume that  $C_{\sup}:=\sup_{X,Z}\left\|\nabla_{\theta}\log p(X| Z) p(Z) \right\|_{\infty}$ is finite, each of the three terms of $\nabla_{\theta}\Delta \Lcalhat_{2^{\ell}}$ is bounded by
\[ \|\nabla_{\theta}\Lcalhat_{2^{\ell}}\|^2_2, \|\nabla_{\theta}\Lcalhat^{(a)}_{2^{\ell-1}}\|^2_2, \|\nabla_{\theta}\Lcalhat^{(b)}_{2^{\ell-1}}\|^2_2\leq 2\card(\theta)C_{\sup}^2,\]
where $\card(\theta)$ denotes the cardinality of $\theta$. Thus it follows from Jensen's inequality that
\[ \|\nabla_{\theta}\Delta \Lcalhat_{2^{\ell}}\|^2_2\leq 2\|\nabla_{\theta}\Lcalhat_{2^{\ell}}\|^2_2+\|\nabla_{\theta}\Lcalhat^{(a)}_{2^{\ell-1}}\|^2_2+ \|\nabla_{\theta}\Lcalhat^{(b)}_{2^{\ell-1}}\|^2_2\leq 8\card(\theta)C_{\sup}^2,\]
so that we have
\[ \EE\left[ \|\nabla_{\theta}\Delta \Lcalhat_{2^{\ell}}\|^2_2 \bsone_A\right] \leq 8\card(\theta)C_{\sup}^2\EE\left[ \bsone_A\right]=8\card(\theta)C_{\sup}^2\PP\left[ A\right].\]
The probability of the event $A$, $\PP\left[ A\right]$, can be bounded by applying Markov inequality and then the Marcinkiewicz-Zygmund inequality as
\begin{align*}
    \PP\left[ A\right] & \leq \PP\left[ \left|\frac{\left[\nabla_{\theta}\Lcalhat^{(a)}_{2^{\ell-1}}\right]_D}{p(X)}-1 \right|>\frac{1}{2}\right]+\PP\left[ \left|\frac{\left[\nabla_{\theta}\Lcalhat^{(b)}_{2^{\ell-1}}\right]_D}{p(X)}-1 \right|>\frac{1}{2}\right] \\
    & = \PP\left[ \left|\frac{\left[\nabla_{\theta}\Lcalhat^{(a)}_{2^{\ell-1}}\right]_D}{p(X)}-1 \right|^s>\frac{1}{2^s}\right]+\PP\left[ \left|\frac{\left[\nabla_{\theta}\Lcalhat^{(b)}_{2^{\ell-1}}\right]_D}{p(X)}-1 \right|^s>\frac{1}{2^s}\right] \\
    & \leq 2^s \EE\left[ \left|\frac{\left[\nabla_{\theta}\Lcalhat^{(a)}_{2^{\ell-1}}\right]_D}{p(X)}-1 \right|^s\right]+2^s \EE\left[ \left|\frac{\left[\nabla_{\theta}\Lcalhat^{(b)}_{2^{\ell-1}}\right]_D}{p(X)}-1 \right|^s\right] \\
    & \leq \frac{2^{s+1}C_{s}}{2^{(\ell-1)s/2 }}\EE_{X,Z}\left[\left| \frac{p(X|Z)p(Z)}{p(X)q(Z; X)}-1\right|^{s}\right],
\end{align*}
with some positive constant $C_s$. As the last expectation is assumed finite in the theorem, we see that the first term on the right-hand side of \eqref{eq:cases} is of order $2^{-\ell s/2}$.

We move on to the second term on the right-hand side of \eqref{eq:cases}, where the following always holds
\begin{align}\label{eq:comp_event} \left|\frac{\left[\nabla_{\theta}\Lcalhat^{(a)}_{2^{\ell-1}}\right]_D}{p(X)}-1 \right|,  \left|\frac{\left[\nabla_{\theta}\Lcalhat^{(b)}_{2^{\ell-1}}\right]_D}{p(X)}-1 \right|, \left|\frac{\left[\nabla_{\theta}\Lcalhat_{2^{\ell}}\right]_D}{p(X)}-1 \right| \leq \frac{1}{2}. \end{align}
Using \eqref{eq:coupling2}, we obtain an equality
\begin{align*}
    \nabla_{\theta}\Delta \Lcalhat_{2^{\ell}} & = \frac{\left[\nabla_{\theta}\Lcalhat_{2^{\ell}}\right]_N}{\left[\nabla_{\theta}\Lcalhat_{2^{\ell}}\right]_D}-\frac{1}{2}\left( \frac{\left[\nabla_{\theta}\Lcalhat^{(a)}_{2^{\ell-1}}\right]_N}{\left[\nabla_{\theta}\Lcalhat^{(a)}_{2^{\ell-1}}\right]_D}+\frac{\left[\nabla_{\theta}\Lcalhat^{(b)}_{2^{\ell-1}}\right]_N}{\left[\nabla_{\theta}\Lcalhat^{(b)}_{2^{\ell-1}}\right]_D}\right) \\
    & = \left( \frac{1}{\left[\nabla_{\theta}\Lcalhat_{2^{\ell}}\right]_D}-\frac{1}{p(X)}\right)\left( \left[\nabla_{\theta}\Lcalhat_{2^{\ell}}\right]_N-\nabla_{\theta} p(X)\right)+\frac{\nabla_{\theta} p(X)}{\left[\nabla_{\theta}\Lcalhat_{2^{\ell}}\right]_D}\left( \frac{\left[\nabla_{\theta}\Lcalhat_{2^{\ell}}\right]_D}{p(X)}-1\right)^2\\
    & \quad - \frac{1}{2}\left( \frac{1}{\left[\nabla_{\theta}\Lcalhat^{(a)}_{2^{\ell-1}}\right]_D}-\frac{1}{p(X)}\right)\left( \left[\nabla_{\theta}\Lcalhat^{(a)}_{2^{\ell-1}}\right]_N-\nabla_{\theta} p(X)\right)-\frac{1}{2}\frac{\nabla_{\theta} p(X)}{\left[\nabla_{\theta}\Lcalhat^{(a)}_{2^{\ell-1}}\right]_D}\left( \frac{\left[\nabla_{\theta}\Lcalhat^{(a)}_{2^{\ell-1}}\right]_D}{p(X)}-1\right)^2\\
    & \quad - \frac{1}{2}\left( \frac{1}{\left[\nabla_{\theta}\Lcalhat^{(b)}_{2^{\ell-1}}\right]_D}-\frac{1}{p(X)}\right)\left( \left[\nabla_{\theta}\Lcalhat^{(b)}_{2^{\ell-1}}\right]_N-\nabla_{\theta} p(X)\right)-\frac{1}{2}\frac{\nabla_{\theta} p(X)}{\left[\nabla_{\theta}\Lcalhat^{(b)}_{2^{\ell-1}}\right]_D}\left( \frac{\left[\nabla_{\theta}\Lcalhat^{(b)}_{2^{\ell-1}}\right]_D}{p(X)}-1\right)^2.
\end{align*}
Applying Jensen's inequality and the inequality \eqref{eq:comp_event} leads to
\begin{align*}
    \|\nabla_{\theta}\Delta \Lcalhat_{2^{\ell}}\|_2^2 & \leq 4\left( \frac{\left[\nabla_{\theta}\Lcalhat_{2^{\ell}}\right]_D}{p(X)}-1\right)^2\left\|\frac{ \left[\nabla_{\theta}\Lcalhat_{2^{\ell}}\right]_N-\nabla_{\theta} p(X)}{\left[\nabla_{\theta}\Lcalhat_{2^{\ell}}\right]_D}\right\|_2^2+4\frac{\|\nabla_{\theta} p(X)\|^2_2}{\left(\left[\nabla_{\theta}\Lcalhat_{2^{\ell}}\right]_D\right)^2}\left( \frac{\left[\nabla_{\theta}\Lcalhat_{2^{\ell}}\right]_D}{p(X)}-1\right)^4\\
    & \quad +2\left( \frac{\left[\nabla_{\theta}\Lcalhat^{(a)}_{2^{\ell-1}}\right]_D}{p(X)}-1\right)^2\left\| \frac{\left[\nabla_{\theta}\Lcalhat^{(a)}_{2^{\ell-1}}\right]_N-\nabla_{\theta} p(X)}{\left[\nabla_{\theta}\Lcalhat^{(a)}_{2^{\ell-1}}\right]_D}\right\|_2^2+2\frac{\|\nabla_{\theta} p(X)\|_2^2}{\left(\left[\nabla_{\theta}\Lcalhat^{(a)}_{2^{\ell-1}}\right]_D\right)^2}\left( \frac{\left[\nabla_{\theta}\Lcalhat^{(a)}_{2^{\ell-1}}\right]_D}{p(X)}-1\right)^4\\
    & \quad +2\left( \frac{\left[\nabla_{\theta}\Lcalhat^{(b)}_{2^{\ell-1}}\right]_D}{p(X)}-1\right)^2\left\| \frac{\left[\nabla_{\theta}\Lcalhat^{(b)}_{2^{\ell-1}}\right]_N-\nabla_{\theta} p(X)}{\left[\nabla_{\theta}\Lcalhat^{(b)}_{2^{\ell-1}}\right]_D}\right\|_2^2+2\frac{\|\nabla_{\theta} p(X)\|_2^2}{\left(\left[\nabla_{\theta}\Lcalhat^{(b)}_{2^{\ell-1}}\right]_D\right)^2}\left( \frac{\left[\nabla_{\theta}\Lcalhat^{(b)}_{2^{\ell-1}}\right]_D}{p(X)}-1\right)^4 \\
    & \leq 16\left( \frac{\left[\nabla_{\theta}\Lcalhat_{2^{\ell}}\right]_D}{p(X)}-1\right)^2\left\|\frac{ \left[\nabla_{\theta}\Lcalhat_{2^{\ell}}\right]_N-\nabla_{\theta} p(X)}{p(X)}\right\|_2^2+16\frac{\|\nabla_{\theta} p(X)\|^2_2}{\left(p(X)\right)^2}\left( \frac{\left[\nabla_{\theta}\Lcalhat_{2^{\ell}}\right]_D}{p(X)}-1\right)^4\\
    & \quad +8\left( \frac{\left[\nabla_{\theta}\Lcalhat^{(a)}_{2^{\ell-1}}\right]_D}{p(X)}-1\right)^2\left\| \frac{\left[\nabla_{\theta}\Lcalhat^{(a)}_{2^{\ell-1}}\right]_N-\nabla_{\theta} p(X)}{p(X)}\right\|_2^2+8\frac{\|\nabla_{\theta} p(X)\|_2^2}{\left(p(X)\right)^2}\left( \frac{\left[\nabla_{\theta}\Lcalhat^{(a)}_{2^{\ell-1}}\right]_D}{p(X)}-1\right)^4\\
    & \quad +8\left( \frac{\left[\nabla_{\theta}\Lcalhat^{(b)}_{2^{\ell-1}}\right]_D}{p(X)}-1\right)^2\left\| \frac{\left[\nabla_{\theta}\Lcalhat^{(b)}_{2^{\ell-1}}\right]_N-\nabla_{\theta} p(X)}{p(X)}\right\|_2^2+8\frac{\|\nabla_{\theta} p(X)\|_2^2}{\left(p(X)\right)^2}\left( \frac{\left[\nabla_{\theta}\Lcalhat^{(b)}_{2^{\ell-1}}\right]_D}{p(X)}-1\right)^4.
\end{align*}
We close the proof by showing that the expectations of the first and second terms on the right-most side above are of order $2^{-\ell \min(s/2,2)}$, respectively, as the other terms can be bounded in the same way.

For the first term, it follows from H\"{o}lder's inequality with the exponents $\min(s/2,2)$ and its conjugate, Jensen's inequality, and the Marcinkiewicz-Zygmund inequality that
\begin{align*}
    & \EE\left[ \left( \frac{\left[\nabla_{\theta}\Lcalhat_{2^{\ell}}\right]_D}{p(X)}-1\right)^2\left\|\frac{ \left[\nabla_{\theta}\Lcalhat_{2^{\ell}}\right]_N-\nabla_{\theta} p(X)}{p(X)}\right\|_2^2 \bsone_{A^c}\right] \\
    & \leq 2^{\max(4-s,0)}\EE\left[ \left( \frac{\left[\nabla_{\theta}\Lcalhat_{2^{\ell}}\right]_D}{p(X)}-1\right)^{\min(s,4)-2}\left\|\frac{ \left[\nabla_{\theta}\Lcalhat_{2^{\ell}}\right]_N-\nabla_{\theta} p(X)}{p(X)}\right\|_2^2 \right] \\
    & \leq 2^{\max(4-s,0)}\left(\EE\left[ \left( \frac{\left[\nabla_{\theta}\Lcalhat_{2^{\ell}}\right]_D}{p(X)}-1\right)^{\min(s,4)}\right]\right)^{1-2/\min(s,4)}\left(\EE\left[\left\|\frac{ \left[\nabla_{\theta}\Lcalhat_{2^{\ell}}\right]_N-\nabla_{\theta} p(X)}{p(X)}\right\|_2^{\min(s,4)} \right]\right)^{2/\min(s,4)} \\
    & \leq 2^{\max(4-s,0)}\left(\frac{C_{\min(s,4)}}{2^{\ell \min(s/2,2)}}\EE\left[ \left( \frac{p(X|Z)p(Z)}{p(X)q(Z; X)}-1\right)^{\min(s,4)}\right]\right)^{1-2/\min(s,4)}\\
    & \quad \times \left(\left(\card(\theta)\right)^{\min(s/2,2)-1}\EE\left[\left\|\frac{ \left[\nabla_{\theta}\Lcalhat_{2^{\ell}}\right]_N-\nabla_{\theta} p(X)}{p(X)}\right\|_{\min(s,4)}^{\min(s,4)} \right]\right)^{2/\min(s,4)} \\
    & \leq 2^{\max(4-s,0)}\left(\frac{C_{\min(s,4)}}{2^{\ell \min(s/2,2)}}\EE\left[ \left( \frac{p(X|Z)p(Z)}{p(X)q(Z; X)}-1\right)^{\min(s,4)}\right]\right)^{1-2/\min(s,4)}\\
    & \quad \times \left(\left(\card(\theta)\right)^{\min(s/2,2)-1}\frac{C_{\min(s,4)}}{2^{\ell\min(s/2,2)}}\EE\left[\left\| \frac{\nabla_{\theta}p(X|Z)p(Z)}{p(X)q(Z; X)}-\frac{\nabla_{\theta} p(X)}{p(X)}\right\|_{\min(s,4)}^{\min(s,4)} \right]\right)^{2/\min(s,4)} \\
    & = 2^{\max(4-s,0)}\left(\card(\theta)\right)^{1-2/\min(s,4)}\frac{C_{\min(s,4)}}{2^{\ell \min(s/2,2)}}\left(\EE\left[ \left( \frac{p(X|Z)p(Z)}{p(X)q(Z; X)}-1\right)^{\min(s,4)}\right]\right)^{1-2/\min(s,4)} \\
    & \quad \times \left(\EE\left[\left\| \frac{\nabla_{\theta}p(X|Z)p(Z)}{p(X)q(Z; X)}-\frac{\nabla_{\theta} p(X)}{p(X)}\right\|_{\min(s,4)}^{\min(s,4)} \right]\right)^{2/\min(s,4)}.
\end{align*}
The two expectations appearing in the right-most side can be shown to be finite by the assumptions given in the theorem. This way we see that the expectation of the first term decays with an order $2^{-\ell \min(s/2,2)}$.

For the second term, a similar argument goes through as
\begin{align*}
    \EE\left[\frac{\|\nabla_{\theta} p(X)\|^2_2}{\left(p(X)\right)^2}\left( \frac{\left[\nabla_{\theta}\Lcalhat_{2^{\ell}}\right]_D}{p(X)}-1\right)^4\bsone_{A^c}\right] & \leq 2^{\max(4-s,0)}\EE\left[\frac{\|\nabla_{\theta} p(X)\|^2_2}{\left(p(X)\right)^2}\left( \frac{\left[\nabla_{\theta}\Lcalhat_{2^{\ell}}\right]_D}{p(X)}-1\right)^{\min(s,4)}\right] \\
    & \leq 2^{\max(4-s,0)}\card(\theta) C_{\sup}^2\EE\left[\left( \frac{\left[\nabla_{\theta}\Lcalhat_{2^{\ell}}\right]_D}{p(X)}-1\right)^{\min(s,4)}\right] \\
    & \leq 2^{\max(4-s,0)}\card(\theta) C_{\sup}^2\frac{C_{\min(s,4)}}{2^{\ell\min(s/2,2)}}\EE\left[\left( \frac{p(X|Z)p(Z)}{p(X)q(Z; X)}-1\right)^{\min(s,4)}\right].
\end{align*}
Since the last expectation is assumed finite, we see that the expectation of the second term decays with an order $2^{-\ell \min(s/2,2)}$. 
\end{proof}

\section{Bayesian Treatment of Parameters}\label{app:lmelbo}

\subsection{Model Settings}
\begin{figure}[!htp]
\centering
\begin{tikzpicture}
  \node[obs]  (x) {$\bsx_n$};
  \node[latent, left=of x]  (z) {$\bsz_n$};
  \node[latent, above=5mm of z]  (w) {$\bstheta$};
  
  \edge {z} {x} ; %
  \edge {w} {x, z} ; %

  \plate {xz} {(x)(z)} {$n = 1,...,N$} ;
  
\end{tikzpicture}
\caption{A graphical model with local and global latent variables.}
\label{fig:graphical_model_2}
\end{figure}
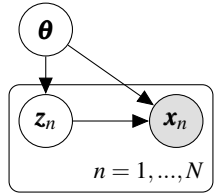

To treat the parameter $\theta$ in a Bayesian manner, we introduce a probabilistic model described by the following data generating process:
\begin{equation*}
    \begin{array}{ll}
    \bstheta &\sim p(\theta)\\
    \bsz_n|\bstheta=\theta &\sim p(z|\theta)\\
    \bsx_n|\bsz_n=z_n, \bstheta=\theta &\sim p(x|z_n, \theta).\\
    \end{array}
\end{equation*}
This model can be also expressed as the graphical model in Figure \ref{fig:graphical_model_2}, wherein $\bstheta$ is treated as a random variable.

\subsection{Locally Marginalized Variational Inference}

For the estimation of this model, we use the variational inference \citep{jordan1999introduction}. In the variational inference, the evidence lower bound (ELBO) for a latent variable model $p_\theta(x, z) = p_\theta(x|z)p_\theta(z) $ plays a central role. It is defined by
$$\text{ELBO} = \log p_\theta(x) - \KL[q_\phi(z)||p_\theta(z|x)],$$ 
where $\KL[ \cdot || \cdot ]$ stands for the Kullback-Leibler (KL) divergence. 
This quantity is maximized with respect to the model parameter $\theta$ as well as the parameter $\phi$ of the variational posterior. 

In our probabilistic model described above, the standard ELBO becomes
$$\text{ELBO}=\log p(x_{1:N}) - \KL[q(z_{1:N},\theta)||p(z_{1:N},\theta|x_{1:N})].$$
However, we can obtain a tighter lower bound on the model evidence by marginalizing out the local variables $\bsz_n$'s. We call our lower bound the locally marginalized ELBO (LMELBO) here, and define it as follows: 
$$\text{LMELBO} :=\log p(x_{1:N}) - \KL[q(\theta)||p(\theta|x_{1:N})].$$ 
Let us consider the mean-field assumption $q(z_{1:N},\theta)=q(\theta)\prod_{n=1}^{N}q(z_n)$ of the variational posterior distribution. By definition, our lower bound on the evidence is tighter than the ELBO due to the following property of the KL divergence:
\begin{align*}
\KL[q(\theta)q(z_{1:N})||p(\theta, z_{1:N}|x_{1:N})] & = \KL[q(\theta)||p(\theta|x_{1:N})] + \EE_{\bstheta\sim q}\KL[q(z_{1:N})||p(z_{1:N}|\bstheta,x_{1:N})] \\
& \geq \KL[q(\theta)||p(\theta|x_{1:N})],    
\end{align*}
where the inequality follows from the non-negativity of the KL divergence.
As the LMELBO is a tighter lower bound than the normal ELBO, this approximates the evidence better and can be used to estimate the variational posterior of $\bstheta$ with smaller bias in the stochastic variational inference settings \citep{hoffman2013stochastic}.  To obtain the objective for the variational inference, we can rewrite our lower bound as a nested expectation as:

\begin{align*}
\text{LMELBO} 
&=\log p(x_{1:N}) - \KL[q(\theta)||p(\theta|x_{1:N})] \\
&= \EE_{\bstheta\sim q(\theta)}\left[\log \frac{\prod_{n=1}^N p(x_n|\bstheta)p(\bstheta)}{q(\bstheta)}\right] \\
&= \sum_{n=1}^N \EE_{\bstheta\sim q(\theta)}\left[  \log \EE_{\bsz_n\sim q(z;x_n)}\left[\frac{p(x_n, \bsz_n|\bstheta)}{q(\bsz_n;x_n)}\right]\right]
  + \EE_{\bstheta\sim q(\theta)}\left[ \log \frac{p(\bstheta)}{q(\bstheta)}\right] \\
&= N \cdot \EE_X \EE_{\bstheta\sim q(\theta)}\left[ \log \EE_{Z\sim q(z;X)}\left[\frac{p(X, Z|\bstheta)}{q(Z;X)}\right] \right]
+ \EE_{\bstheta\sim q(\theta)}\left[ \log \frac{p(\bstheta)}{q(\bstheta)}\right],
\end{align*}
where we recall that $X$ is a random variable taking $x_1\ldots,x_N$ uniformly. Here, an outer expectation is taken with respect to both $X$ and $\bstheta$ simultaneously, whereas an inner conditional expectation is with respect to $Z$.

\subsection{Application of MLMC}
Applying the MLMC method to estimate the LMELBO is straightforward; we just have to change the definition of $\Lcalhat_{M,K}$ to the following:
\begin{equation*}
    \Lcalhat_{M,K} = \frac{N}{M}\sum_{m=1}^M \log\left[\frac{1}{K}\sum_{k=1}^{K}\frac{p(X_m,Z_{m, k}|\Theta_m)}{q(Z_{m,k};X_{m, k})}\right] - \EE_{\bstheta \sim q(\theta)}\left[\log\frac{p(\bstheta)}{q(\bstheta)}\right].
\end{equation*}
Here, for each $m$, $X_m$'s are uniformly random samples taken from the data $x_1, ..., x_N$ and $\Theta_m$'s are taken from $q(\theta)$. The inner Monte Carlo samples $Z_{m,1}, ..., Z_{m, K}$ are i.i.d.\ random samples from a proposal distribution $q(z_m; X_m)$. The second term, which corresponds to the KL divergence between $q(\theta)$ and $p(\theta)$, can be computed independently from the first term, to which MLMC is applied. In some cases, the KL term can be calculated analytically if we choose $q(\theta)$ and $p(\theta)$ appropriately, e.g., when we use Gaussian for both, the KL divergence between them has an analytical form.

\subsection{Theoretical Results for MLMC}
The theorems for the coupled correction terms for the MLMC estimator can be obtained for the LMELBO after minor modifications to Theorems~\ref{thm:evidence} and \ref{thm:gradient} of the main article:

\begin{theorem}\label{thm:evidence2}
If there exist $s,t>2$ with $(s-2)(t-2)\geq 4$ such that
\begin{align*} 
& \EE_{X}\EE_{\Theta\sim q(\theta)}\left[ \int \left|\frac{p(X, Z| \Theta) }{p(X|\Theta)q(Z;X)}\right|^s\rd Z\right]<\infty, \quad\text{and}\\
& \EE_{X}\EE_{\Theta\sim q(\theta)}\left[ \int \left|\log \frac{p(X, Z| \Theta) }{p(X|\Theta)q(Z;X)}\right|^t\rd Z\right]<\infty,
\end{align*}
the MLMC estimator for the LMELBO satisfies
\[ \alpha = \min\left\{\frac{s(t-1)}{2t}, 1 \right\}\; \; \text{and}\; \; \beta=\min\left\{\frac{s(t-2)}{2t}, 2\right\}, \]
where the parameters $\alpha$ and $\beta$ in the last line indicate the positive constants in Theorem~\ref{thm:mlmc} of the main article.
\end{theorem}

\begin{theorem}\label{thm:gradient2}
If there exists $s\geq 2$ such that
\begin{align*} 
& \EE_{X}\EE_{\Theta\sim q(\theta)}\left[ \int \left|\frac{p(X, Z|\Theta) }{p(X|\Theta)q(Z;X)}\right|^s\rd Z\right]<\infty, \quad\text{and}\\
& \sup_{x,z,\theta}\left\|\nabla_{\theta}\log p(x, z|\theta) \right\|_{\infty}<\infty,
\end{align*}
the MLMC estimator for the gradient of the model evidence satisfies
\[ \alpha = \min\{s/2, 1\} \quad\text{and}\quad \beta = \min\{ s/2, 2\}, \]
where the parameters $\alpha$ and $\beta$ in the last line indicate the positive constants in Theorem~\ref{thm:mlmc} of the main article.
\end{theorem}

Here, the only difference from the theorems for the model evidence is that the outer expectations or the supremum are taken with respect not only to $x$ and $z$ but also to the global parameter $\theta$.

\section{Experiments using Locally Marginalised ELBO}\label{app:lmelbo_exp}

To confirm our MLMC approach can be efficiently applied to the locally marginalized variational inference, we checked the convergence order of the coupled correction term of the Bayesian random effect logistic regression model. 
Furthermore, we examined a Gaussian process classification model combined with LMELBO. In the latter experiment, we observed that the MLMC approach actually works well in a real-world example.

\subsection{Bayesian Random Effect Logistic Regression}
To treat the coefficients $\w_0$ and $\w$ in a Bayesian manner, we set the following prior distribution to them:
\begin{align*}
    \bsw_0 &\sim N(0, 1) \\
    \bsw &\sim N(0, I_D)
\end{align*}
Then, conditionally on $\bsw_0$ and $\bsw$, Bayesian random effect logistic regression has the i.i.d.\ data generating process identical to the non-Bayesian model for $n = 1, 2, \ldots, N$ and $t=1, \ldots, T$:
\begin{align*}
    \bsz_n &\sim N(0,\tau^2) \\
    \bsy_{n,t} &\sim \text{Bernoulli}\left(p_n\right),
\end{align*}
where we set the logit $p_n$ to $p_n = \sigma(\bsz_n + \bsw_0 + \bsw^T x_{n,t})$.

Again, in the experiment, we used a synthetic data generated from a model whose parameters are given by $\eta=1.0$, $\w_0=0$, $\w=(0.25, 0.50, 0.75)^T$. To keep the parameter $\tau^2$ positive, we parametrized it with a non-constrained parameter $\eta$ by the softplus transformation as $\tau^2=\log{\left(1+\exp(\eta)\right)}$. 

\begin{figure}[ht]
    \centering
    \begin{subfigure}{0.47\textwidth}
    \includegraphics[width=\linewidth]{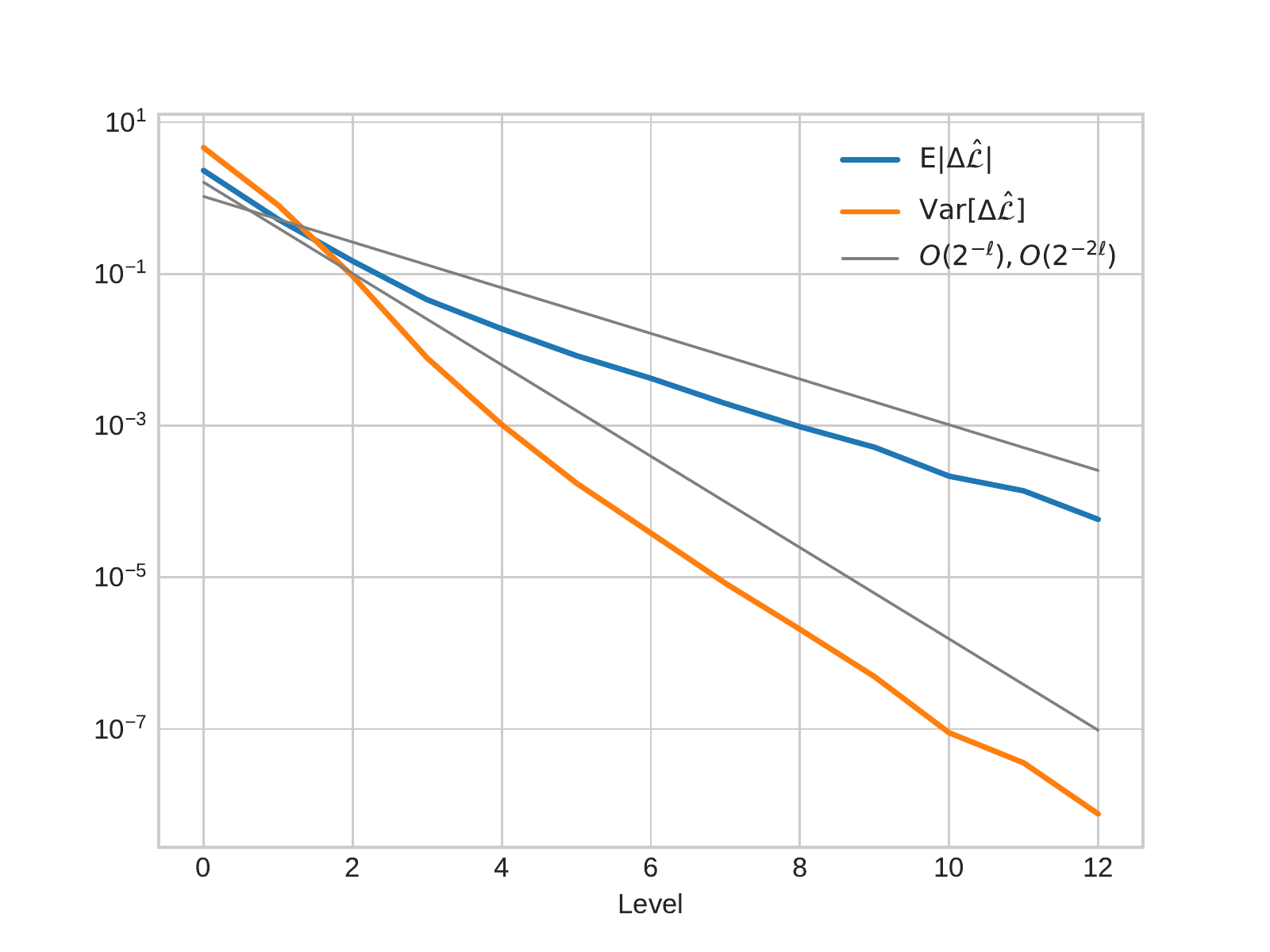}
    \caption{Decay of $\Delta\Lcalhat_{1,2^{\ell}}$}
    \label{fig:conv-delta2} 
    \end{subfigure}
    \begin{subfigure}{0.47\textwidth}
    \includegraphics[width=\linewidth]{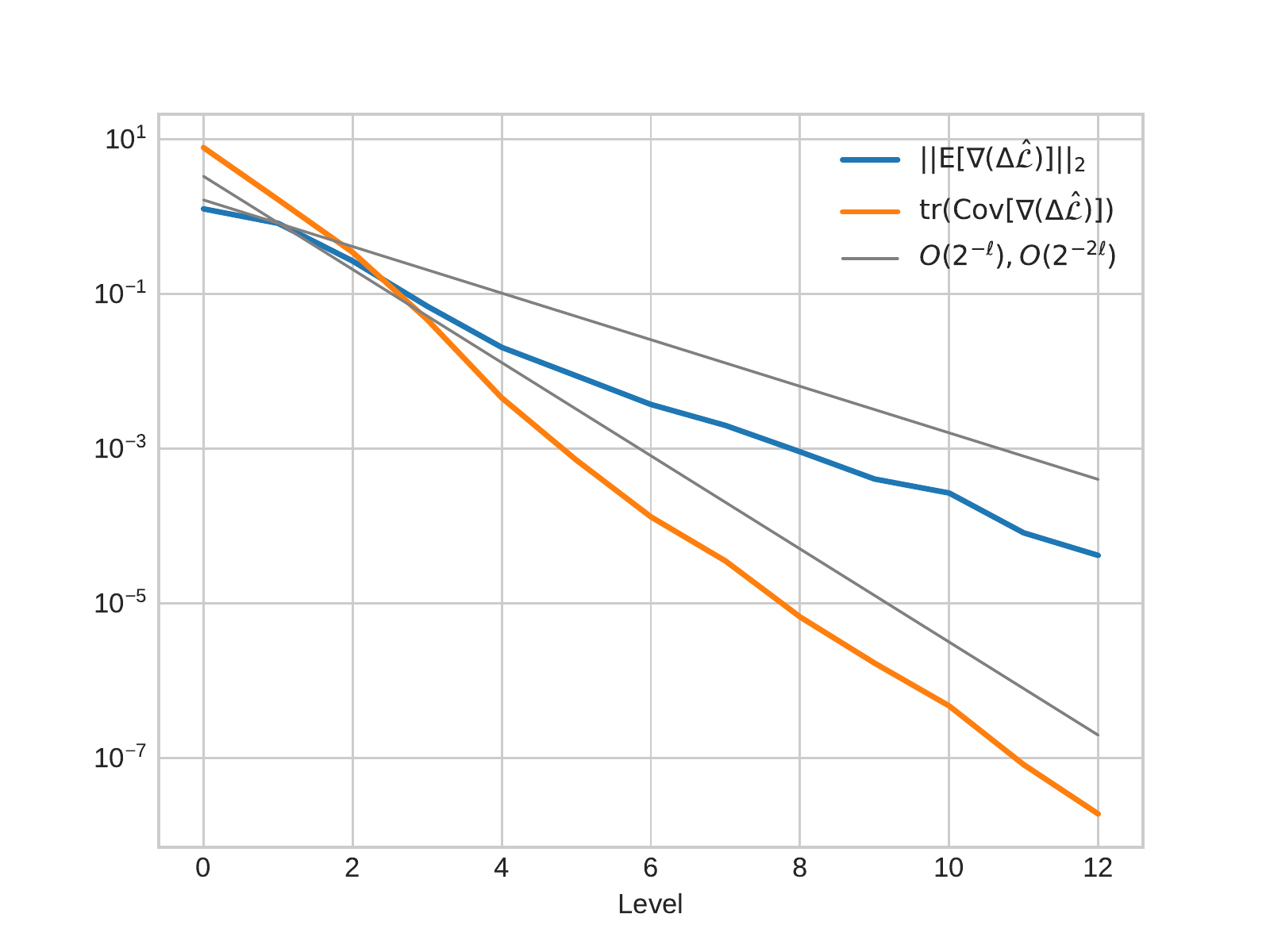}
    \caption{Decay of $\nabla_{\theta}\Delta\Lcalhat_{1,2^{\ell}}$}
    \label{fig:conv-grad-delta2} 
    \end{subfigure}
    \caption{Convergence of the mean and variance of the coupled correction estimators of Bayesian random effect logistic regression.}
\end{figure}

To examine whether the assumptions required for the MLMC estimation in Theorem~\ref{thm:mlmc} are satisfied, we evaluated the convergence behavior of the corrections $\Delta\Lcalhat_{1,2^{\ell}}$ and their gradient counterparts $\nabla_{\theta}\Delta\Lcalhat_{1,2^{\ell}}$.  

Figure~\ref{fig:conv-delta2} and \ref{fig:conv-grad-delta2} shows the convergence behaviors of $\EE[\Delta\Lcalhat_{1,2^{\ell}}]$, $\VV[\Delta\Lcalhat_{1,2^{\ell}}]$ and their gradient counterparts. We see that the expectation and the variance approximately decay with the orders of $2^{-\ell}$ and $2^{-2\ell}$, respectively, implying that we have $\alpha=1$ and $\beta=2$ in the assumptions of Theorem~\ref{thm:mlmc}. 

In Tables~\ref{tab:lmelbo_posterior} and \ref{tab:elbo_posterior}, the posterior estimates obtained by optimizing LMELBO and ELBO are presented. Though the posterior mean estimated with ELBO is biased, the bias for the LMELBO is much smaller. 
Moreover, the true parameters are in the one standard deviation intervals of the estimated posterior when LMELBO is used. This implies that the estimates of the standard deviations are good measures of the uncertainty of the estimation.
\begin{table*}[ht]
\small
\begin{center}
\caption{Posterior of coefficients ($\bsw$) of the Bayesian random effect regression model estimated by the LMELBO with $K=512$.}
\small

\begin{tabular}{l  r  r  r  r } 
\toprule
  & $\bsw_0$  & $\bsw_1$  & $\bsw_2$  & $\bsw_3$ \\
\midrule
Ground Truth & 0.0 & 0.25 & 0.5 & 0.75 \\
Posterior Mean & 0.0069 & 0.2700 & 0.5197 & 0.7764 \\
Posterior Stddev & 0.0351 & 0.0335 & 0.0335 & 0.0348 \\
\bottomrule
\end{tabular}
\label{tab:lmelbo_posterior}
\end{center}
\end{table*}

\begin{table*}[ht]
\small
\begin{center}
\caption{Posterior of coefficients ($\bsw$) of Bayesian random effect regression model estimated by the ELBO, which is equivalent to the LMELBO with $K=1$.}
\small

\begin{tabular}{l  r  r  r  r } 
\toprule
  & $\bsw_0$  & $\bsw_1$  & $\bsw_2$  & $\bsw_3$ \\
\midrule
Ground Truth & 0.0 & 0.25 & 0.5 & 0.75 \\
Posterior Mean & 0.0163 & 0.2356 & 0.4613 & 0.6645 \\
Posterior Stddev & 0.0295 & 0.0293 & 0.0300 & 0.0310 \\
\bottomrule
\end{tabular}
\label{tab:elbo_posterior}
\end{center}
\end{table*}

\subsection{Gaussian Process Classification}
\interdisplaylinepenalty=10000

To see the effectiveness of locally marginalized variational inference, we considered a Gaussian process classification model with fully independent training conditional (FITC) approximation (\citeauthor{snelson2006sparse}, \citeyear{snelson2006sparse} and \citeauthor{hensman2013gaussian}, \citeyear{hensman2013gaussian}), which can be written as 
\begin{align*}
f &\sim\mathrm{GP}_{\text{FITC}} \quad \text{ and}\\
Y_n &\sim \mathrm{Bernoulli}(\sigma(f(x_n))),
\end{align*}
where $\sigma(x)=1/(1 + \exp(-x))$ is the sigmoid link function. Under the FITC approximation, we approximate the Gaussian process evaluated at $x_{1:N}$ with the following joint distribution:
$$ p_{\text{FITC}}(f(x_1), \ldots, f(x_N), f(z_{1:M})) = \prod_{n=1}^N p_{\mathrm{GP}}(f(x_n)|f(z_{1:M}))p_{\mathrm{GP}}(f(z_{1:M})).$$
Here, $z_{1:M}$ is a vector of inducing points. The densities $p_{\mathrm{GP}}(f(x_n)|f(z_{1:M}))$ and $p_{\mathrm{GP}}(f(z_{1:M}))$ denote the conditional distribution of $f(x_n)$ and distribution of $f(z_{1:M})$ induced by the original Gaussian process. Each of $f(x_n)$ is the local latent variable to be marginalized in LMELBO, and $f(z_{1:M})$ is the global latent variable to which variational inference is applied.

In the experiment, we applied the above Gaussian process classification model on the adult dataset \citep{Dua:2019} to predict if the income is more than 50K. We used the automatic relevance detection (ARD) kernel and optimized its parameter along with the parameter of the variational posterior $q(f(x_{1:M}))$. As a conditional proposal distribution $q(f(x_n)|f(z_{1:M}))$ given the global latent variable $f(z_{1:M})$, we used $p_{\mathrm{GP}}(f(x_n)|f(z_{1:M}))$. The MLMC estimator was constructed as $\Lcalhat^{\text{MLMC}}_{16\cdot 2^L} = \Lcalhat_{M_0, 16} + \sum_{\ell=1}^L \Delta \Lcalhat_{M_\ell, 16 \cdot 2^\ell}$, as we found that $\Lcalhat_{1, 16}$ is more numerically stable and has a much smaller variance than $\Lcalhat_{1, 1}$.

\begin{figure}[ht]
    \centering
    \begin{subfigure}{0.47\textwidth}
    \includegraphics[width=\linewidth]{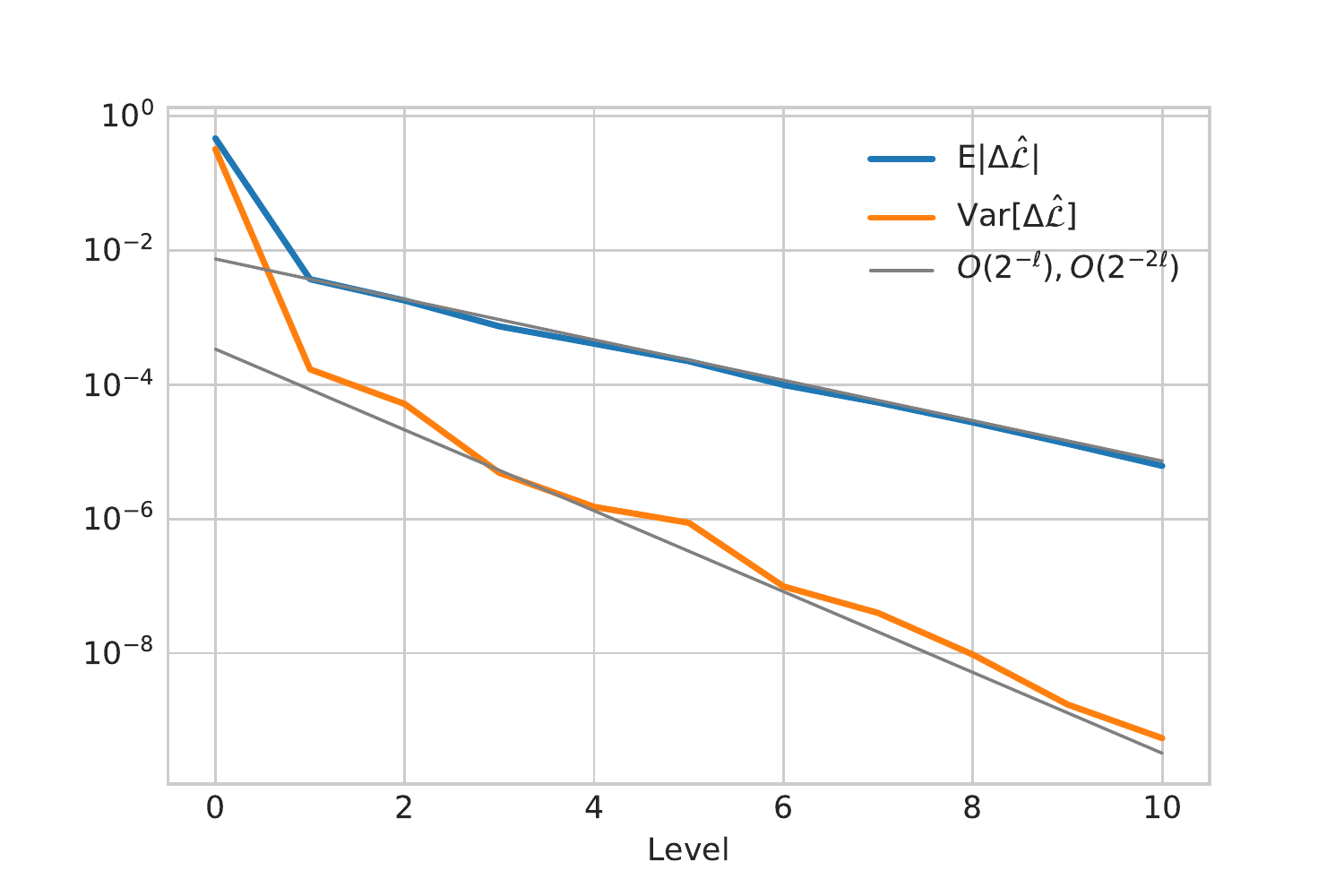}
    \caption{Decay of $\Delta\Lcalhat_{1,16\cdot2^{\ell}}$}
    \label{fig:gpc-conv-delta} 
    \end{subfigure}
    \begin{subfigure}{0.47\textwidth}
    \includegraphics[width=\linewidth]{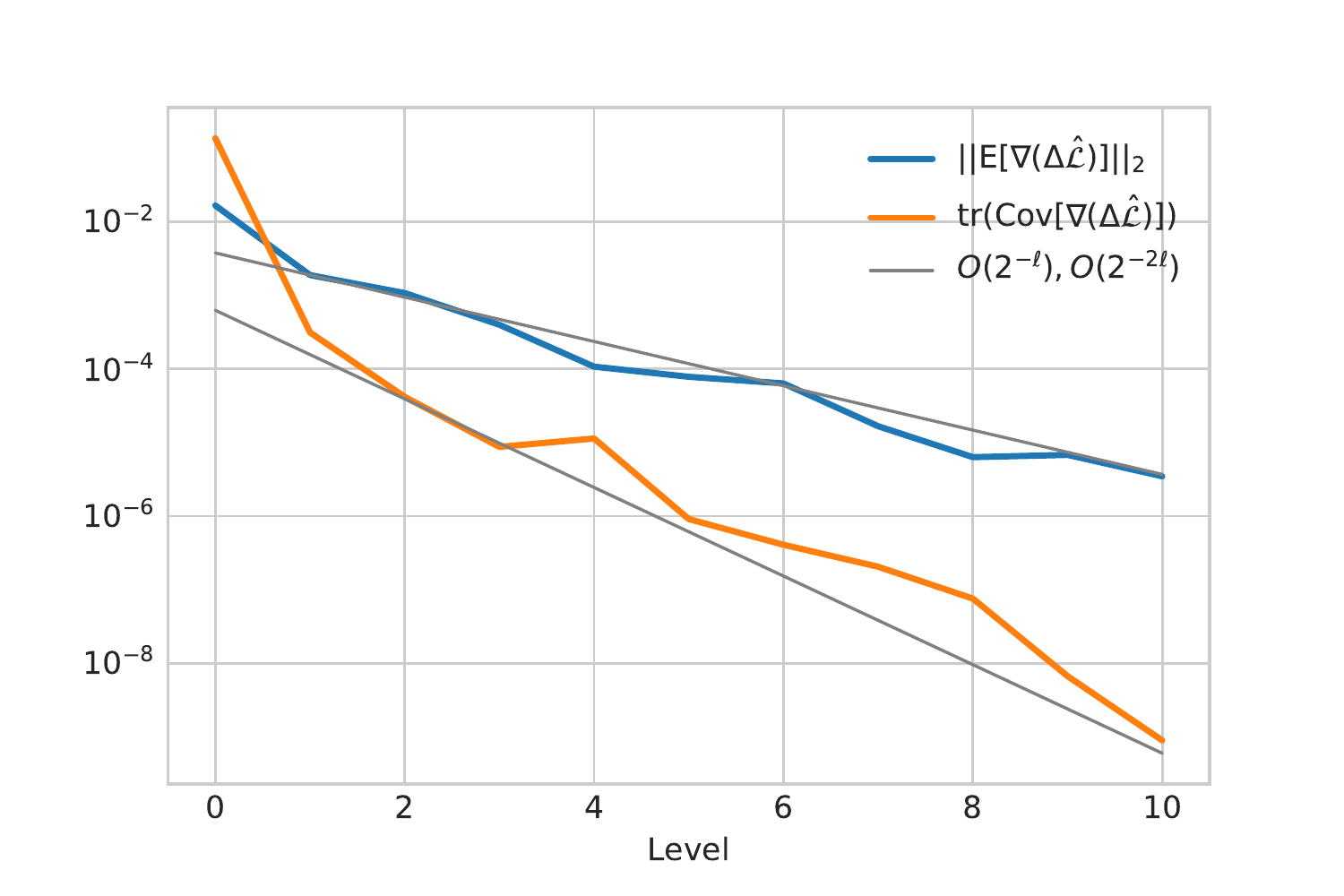}
    \caption{Decay of $\nabla_{\theta}\Delta\Lcalhat_{1,16\cdot2^{\ell}}$}
    \label{fig:gpc-conv-grad-delta} 
    \end{subfigure}
    \caption{Convergence of the mean and variance of the coupled correction estimators of Gaussian process classification with FITC approximation.}
\end{figure}

Figure~\ref{fig:gpc-conv-delta} and \ref{fig:gpc-conv-grad-delta} shows the convergence behaviors of $\EE[\Delta\Lcalhat_{1,16\cdot 2^{\ell}}]$, $\VV[\Delta\Lcalhat_{1,16\cdot 2^{\ell}}]$ and their gradient counterparts on the adult dataset. We again see the exponential decays, confirming that the assumptions for the MLMC method are satisfied. Moreover, we observed that the model trained by LMELBO has a higher average log marginal likelihood for test data (-0.395±0.02), than the model trained by the ELBO (-0.436±0.03). This is as expected from the construction of the objectives, LMELBO and ELBO, as the former is a tighter lower bound of the log marginal likelihood. Here, we approximated the log marginal likelihood by the Monte Carlo average of IWELBO with $k=512$.


\end{document}


%

%

\onecolumn

\LARGE
Moved to the end of the main article due to the submission regulations.